\documentclass[10pt, conference]{IEEEtran}
\usepackage{amsmath,amssymb,amsfonts}
\usepackage{graphicx}
\usepackage{pifont}
\usepackage{tikz}

\usepackage{amsthm}
\usepackage{xcolor}
\usepackage[figuresright]{rotating}

\usepackage{graphicx}
\graphicspath{{/}{fig/}}

\usepackage{array}
\usepackage{textcomp}
\usepackage{xcolor}
\usepackage{multirow}
\usepackage{booktabs}

\usepackage{mathtools}
\usepackage{breqn}
\usepackage{float}

\usepackage{fancyhdr}

\usepackage{pgfplots}
\pgfplotsset{compat=1.7}
\usepgfplotslibrary{groupplots}

\usepackage{silence}
\usepackage{multirow,tabularx}
\usepackage{hyperref}
\usepackage{flushend}
\usepackage{algorithmic}
\usepackage[vlined, ruled, shortend]{algorithm2e}

\usepackage{subcaption}

\newlength\figureheight
\newlength\figurewidth
\SetAlCapNameFnt{\footnotesize}
\SetAlCapFnt{\footnotesize}

\title{
    UAV Tracking with Lidar as a Camera Sensor in GNSS-Denied Environments \\
}

\author{
    \IEEEauthorblockN{
        \vspace{1em}
        Ha Sier\IEEEauthorrefmark{1}\IEEEauthorrefmark{2},
        Xianjia Yu\IEEEauthorrefmark{2},
        Iacopo Catalano\IEEEauthorrefmark{2},
        Jorge Pe\~na Queralta\IEEEauthorrefmark{2},
        Zhuo Zou\IEEEauthorrefmark{1},
        Tomi Westerlund\IEEEauthorrefmark{2}
    }
    \IEEEauthorblockA{
        \normalsize
        \IEEEauthorrefmark{1}School of Information Science and Technology, Fudan Universtiy, China\\
        \IEEEauthorrefmark{2}\href{https://tiers.utu.fi}{Turku Intelligent Embedded and Robotic Systems (TIERS) Lab, University of Turku, Finland}.\\
        Emails: \textsuperscript{1}\{seha20, zhuo\}@fudan.edu.cn, \textsuperscript{2}\{xianjia.yu, imcata, jopequ, tovewe\}@utu.fi\\[+6pt]
    }
}

\fancypagestyle{FirstPage}{

\lfoot{979-8-3503-2308-5/23/\$31.00 ©2023 IEEE}
}

\begin{document}

\maketitle
\thispagestyle{empty}
\pagestyle{empty}

%%%%%%%%%%%%%%%%%%%%%%%%%%%%%%%%%%%%%%%%%%%%%%
%%                                          %%
%%           ABSTRACT AND TITLE             %%
%%                                          %%
%%%%%%%%%%%%%%%%%%%%%%%%%%%%%%%%%%%%%%%%%%%%%%

%%%%%%%%%%%%%%%%%%%%%%%%%%%%%%%%%%%%%%%%%%%%%%
%%                                          %%
%%                ABSTRACT                  %%
%%                                          %%
%%%%%%%%%%%%%%%%%%%%%%%%%%%%%%%%%%%%%%%%%%%%%%

\begin{abstract}%
    \label{sec:abstract}%
    Light detection and ranging (LiDAR) sensor has become one of the primary sensors in robotics and autonomous system for high-accuracy situational awareness. In recent years, multi-modal LiDAR systems emerged, and among them, LiDAR-as-a-camera sensors provide not only 3D point clouds but also fixed-resolution 360\textdegree panoramic images by encoding either depth, reflectivity, or near-infrared light in the image pixels. This potentially brings computer vision capabilities on top of the potential of LiDAR itself. In this paper, we are specifically interested in utilizing LiDARs and LiDAR-generated images for tracking Unmanned Aerial Vehicles (UAVs) in real-time which can benefit applications including docking, remote identification, or counter-UAV systems, among others. This is, to the best of our knowledge, the first work that explores the possibility of fusing the images and point cloud generated by a single LiDAR sensor to track a UAV without a priori known initialized position. We trained a custom YOLOv5 model for detecting UAVs based on the panoramic images collected in an indoor experiment arena with a motion capture (MOCAP) system. By integrating with the point cloud, we are able to continuously provide the position of the UAV. Our experiment demonstrated the effectiveness of the proposed UAV tracking approach compared with methods based only on point clouds or images. Additionally, we evaluated the real-time performance of our approach on the Nvidia Jetson Nano, a popular mobile computing platform. 

\end{abstract}

\begin{IEEEkeywords}

    UAV; LiDAR-as-a-camera; drone tracking; LiDAR;
    LiDAR detection; LiDAR tracking; YOLOV5; 

\end{IEEEkeywords}
\IEEEpeerreviewmaketitle

%%%%%%%%%%%%%%%%%%%%%%%%%%%%%%%%%%%%%%%%%%%%%%
%%                                          %%
%%                SECTIONS                  %%
%%                                          %%
%%%%%%%%%%%%%%%%%%%%%%%%%%%%%%%%%%%%%%%%%%%%%%
%%%%%%%%%%%%%%%%%%%%%%%%%%%%%%%%%%%%%%%%%%%%%%
%%                                          %%
%%              INTRODUCTION                %%
%%                                          %%
%%%%%%%%%%%%%%%%%%%%%%%%%%%%%%%%%%%%%%%%%%%%%%

\section{Introduction}\label{sec:introduction}
\thispagestyle{FirstPage}

Achieving high-level situational awareness is a vital but %formidable 
not trivial task in the domain of mobile robotics and autonomous systems in general, as it enables effective decision-making, navigation, and control in %heterogeneous 
complex environments~\cite{fan2019key}. In recent years, researchers have exploited diverse sensor modalities to enhance robotics perception systems. Among these, LiDAR and cameras have emerged as the primary perception components~\cite{kato2018autoware}. In contrast to cameras, LiDARs offer key features, including long-range and precise geometry data, and generally greater resilience to adverse weather conditions, such as fog and rain. 
Currently, there are various modalities of LiDARs available~\cite{qingqing2022multi}. Ouster LiDARs~\cite{tampuu2022lidar, angus2018lidar, xianjia2022analyzing} stand out by offering dense point clouds and 360\textdegree field of view with low-resolution images, obtained by encoding depth, reflectivity, or near-infrared light in the image pixels. These are so-called lidar as a camera sensors~\cite{tampuu2022lidar, xianjia2022analyzing}. Other lidar manufacturers are expected to provide similar functionality in the near future.

\begin{figure}[t]
    \centering
    \includegraphics[width=0.48\textwidth]{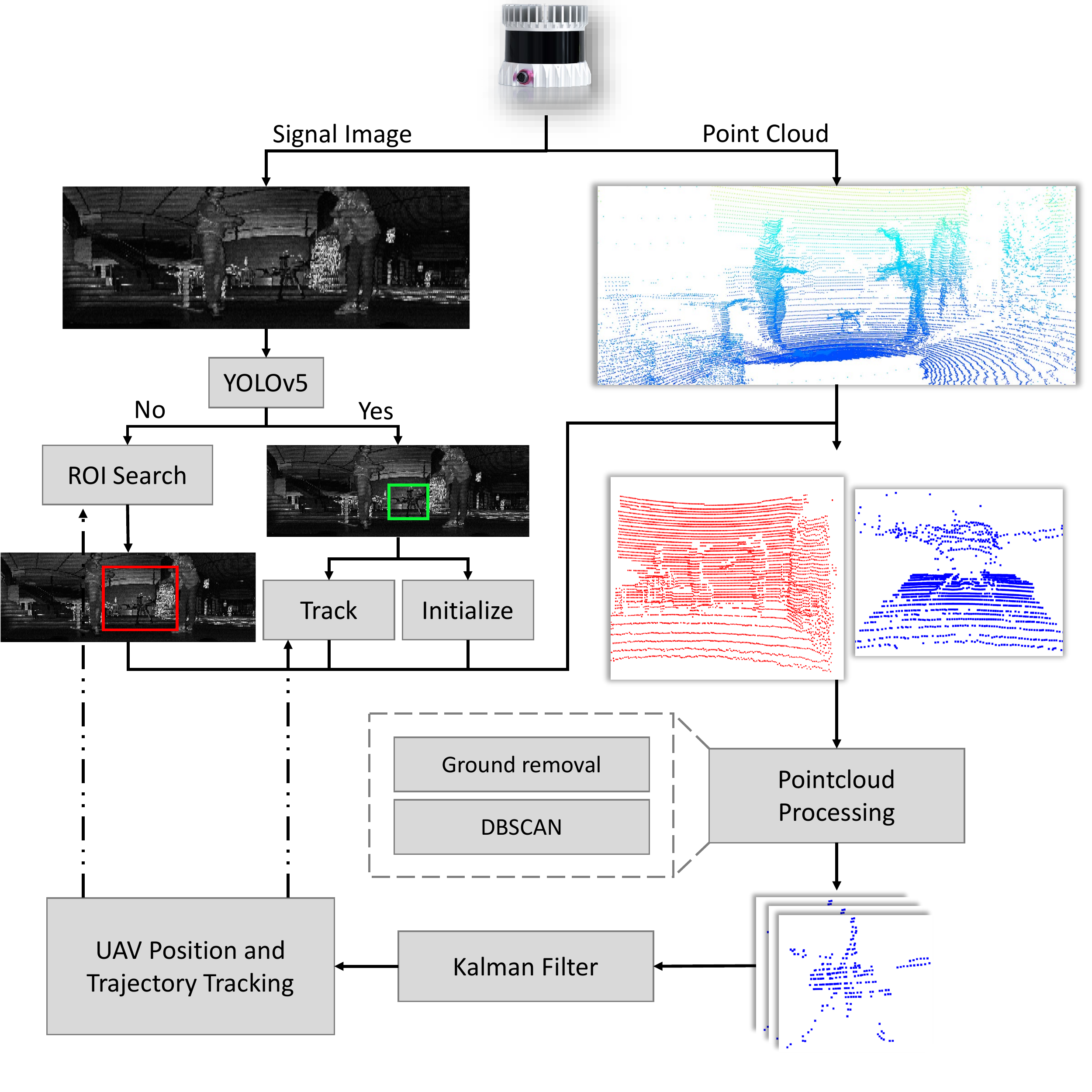}
    \caption{Diagram of proposed UAV tracking system based on the image and point cloud generated by an Ouster LiDAR.}
    \label{fig:concept}
\end{figure}

This type of LiDAR as a camera sensor is inherently compatible with deep learning (DL) models for vision sensors, obviating the need for external camera mounting and calibration~\cite{xianjia2022analyzing}. Moreover, the LiDAR-as-camera approach can augment conventional LiDAR-based methods for object detection and tracking, as the algorithms for the latter are more computationally intensive and less mature than vision sensors. Despite the low vertical resolution and lack of color, the key motivation for considering such an approach is simply to do more with already existing data and without additional sensors.

We are interested in studying the potential of these sensors for tracking unmanned aerial vehicles (UAVs), fusing both lidar point cloud data and signal images. Given that the images generated by the LiDAR as a camera sensors have low resolution (see for example in figure~\ref{fig:pcd_frame}), we are particularly interested in short-range tracking of UAVs, which finds potential applications in UAV docking, or collaboration, among others. The deployment of UAVs requires accurate localization, especially in cases where GNSS signals suffer from degradation or are not available~\cite{li2018high}. 

In this paper, we propose a UAV tracking approach based on the integration of images and 3D point clouds generated by an Ouster LiDAR sensor. The diagram of the approach we follow is illustrated in figure~\ref{fig:concept}. The UAV can be detected in signal images instead of manually giving its initial position as it is needed in other point-cloud-only approaches. The detection also yields an approximate region of interest (ROI) in the point cloud, which will be expanded if no detection occurs. This approach reduces computation overhead by avoiding the need for an overall point cloud search. UAV identification is achieved by clustering points within the ROI, followed by continuous position estimation using the Kalman Filter (KF).

The remainder of this document is organized as follows. Section II introduces the current state-of-the-art UAV tracking methods relevant to the presented approach. Section III  introduces the methodology and the experimental setup with results presented in Section IV. Section V concludes the work and outlines future research directions.

\section{Related Work} \label{sec:related_work}

This section reviews the literature in the field of UAV detection and tracking. Due to the limited research on tracking small objects such as UAVs based on LiDAR, we focus on: (i) vision-based and (ii) LiDAR-based UAV tracking; (iii) research into the potential of LiDAR-as-a-camera sensors; and (iv) applications of UAV tracking.

\subsection{UAV tracking with cameras}

Vision-based methods are widely used to track small objects and UAVs~\cite{mueller2016benchmark, 5354489, 8988144}. They can be divided into two categories: those that rely on passive or active visual markers, and those that detect and track objects in general, e.g., with traditional computer vision or deep learning.
For example, \cite{5354489}~introduces a trinocular system with ground-based cameras to control a rotary-wing UAV in real time based on its key features. Alternatively, \cite{6696776} presents an infrared binocular vision system with PTU and exosensors to track drones cheaply under any weather and time conditions based on their infrared spectra. Recent developments in deep convolutional neural networks (CNNs) have boosted adoption in the field of object detection and tracking. Arguably, a large part of the state-of-the-art in tracking is based on deep learning methods. Recent advances in deep CNNs have improved object detection and tracking performance~\cite{li2018deep}. For instance, \cite{8988144}~proposes a CNN-based markerless UAV relative positioning system that allows the stable formation and autonomous interception of multiple UAVs.

Depth cameras can also detect UAVs and help them avoid obstacles using deep learning models that process depth maps ~\cite{carrio2018drone}. While depth cameras can provide accurate position and size measurements, and vision sensors are generally capable of robust tracking and relative localization, our focus in this paper is on the use of Ouster LiDARs because of their flexibility with respect to environmental conditions and their ability to provides more accurate and informative signal images than depth cameras.

\subsection{UAV tracking with LiDARs}

While LiDAR systems are often employed for detecting and tracking objects, they pose unique challenges in detecting and tracking UAVs due to their small size, varied shapes and materials, high speed, and unpredictable movements.

When deployed from a ground robot, a crucial parameter is relative localization between different devices. Li et al.~\cite{qingqing2021adaptive} suggest a new approach for tracking UAVs using LiDAR point clouds. They take into account the UAV speed and distance to adjust the LiDAR frame integration time, which affects the density and size of the point cloud to be processed.

By conducting a probabilistic analysis of detection and ensuring proper setup, as shown in~\cite{dogru2022dronedetection}, it is possible to achieve detection using fewer LiDAR beams, while performing continuous tracking only on a small number of hits. The limitations in the 3D LiDAR technology can be overcome by moving the sensor to increase the field of view and improve the coverage ratio. Additionally, combining a segmentation approach and a simple object model while leveraging temporal information in~\cite{razlaw2019DetectionAT} has been shown to reduce parametrization effort and generalize to different settings.

Another approach, departing from the typical sequence of track-after-detect, is to leverage motion information by searching for minor 3D details in the 360$^{\circ}$ LiDAR scans of the scene. If these clues persist in consecutive scans, the probability of detecting a UAV increases. Furthermore, analyzing the trajectory of the tracked object enables the classification of UAVs and non-UAV objects by identifying typical movement patterns~\cite{hammer2018potentiallidardetection, hammer2018lidarsmalluavs}.

\subsection{LiDAR as a camera}

The Ouster LiDAR blurs the traditional boundaries between cameras and LiDARs by providing 360\textdegree panoramic images, including depth, signal, and ambient images in the near-infrared spectrum, in addition to the point cloud output. These signal and ambient images capture the strength of laser light reflected back to the sensor and ambient light, respectively. Ouster has proven that training models based on these images are effective for various computer vision applications, such as object detection, segmentation, feature extraction, and optical flow~\cite{tampuu2022lidar,angus2018lidar}. Several existing DL models have also been demonstrated to be effective in analyzing these images~\cite{xianjia2022analyzing}. 

\subsection{Applications of UAV tracking}

Recently, researchers have shown interest in tracking and detecting UAVs due to two primary reasons: the rising demand for identifying and detecting foreign objects or drones in areas with controlled airspace, like airports~\cite{guvenc2018detection, hengy2017multimodal}, and the potential for optimizing the utilization of UAVs as versatile mobile sensing platforms through tracking and detection~\cite{queralta2020autosos}.

The ability to track UAVs from unmanned ground vehicles (UGVs) allows for miniaturization and greater flexibility in multi-robot systems, reducing the need for high-accuracy onboard localization. This was demonstrated in the DARPA Subterranean challenge~\cite{rouvcek2019darpa, petrlik2020robust}, where UAVs were dynamically deployed from UGVs in GNSS-denied environments. Localization and collaborative sensing were key challenges, with reports indicating that LiDAR-based tracking was useful in domains where visual-inertial odometry (VIO) has limitations, such as low-visibility situations~\cite{queralta2020vio, qingqing2019offloading}.

Similarly, tracking UAVs is crucial in the landing phase of the aerial system. Different methods using a ground-based stereo camera~\cite{kong2013uavstereolanding} or having the UAV carry an infrared camera to detect signals from the destination~\cite{gui2013uavinfraredlanding} have been proposed. As these works employ cameras as their main sensory system, they can be easily affected by background lighting conditions while in our approach we prefer a LiDAR which is more resilient in these environmental conditions.

% \newpage
%\input{sec/03_ProblemDefinition}
%%%%%%%%%%%%%%%%%%%%%%%%%%%%%%%%%%%%%%%%%%%%%%
%%                                          %%
%%              METHODOLOGY                 %%
%%                                          %%
%%%%%%%%%%%%%%%%%%%%%%%%%%%%%%%%%%%%%%%%%%%%%%

% \newpage
\section{Methodology}

\subsection{Hardware Information}

\begin{figure}[t] 
    \centering   
    \includegraphics[width=0.48\textwidth]{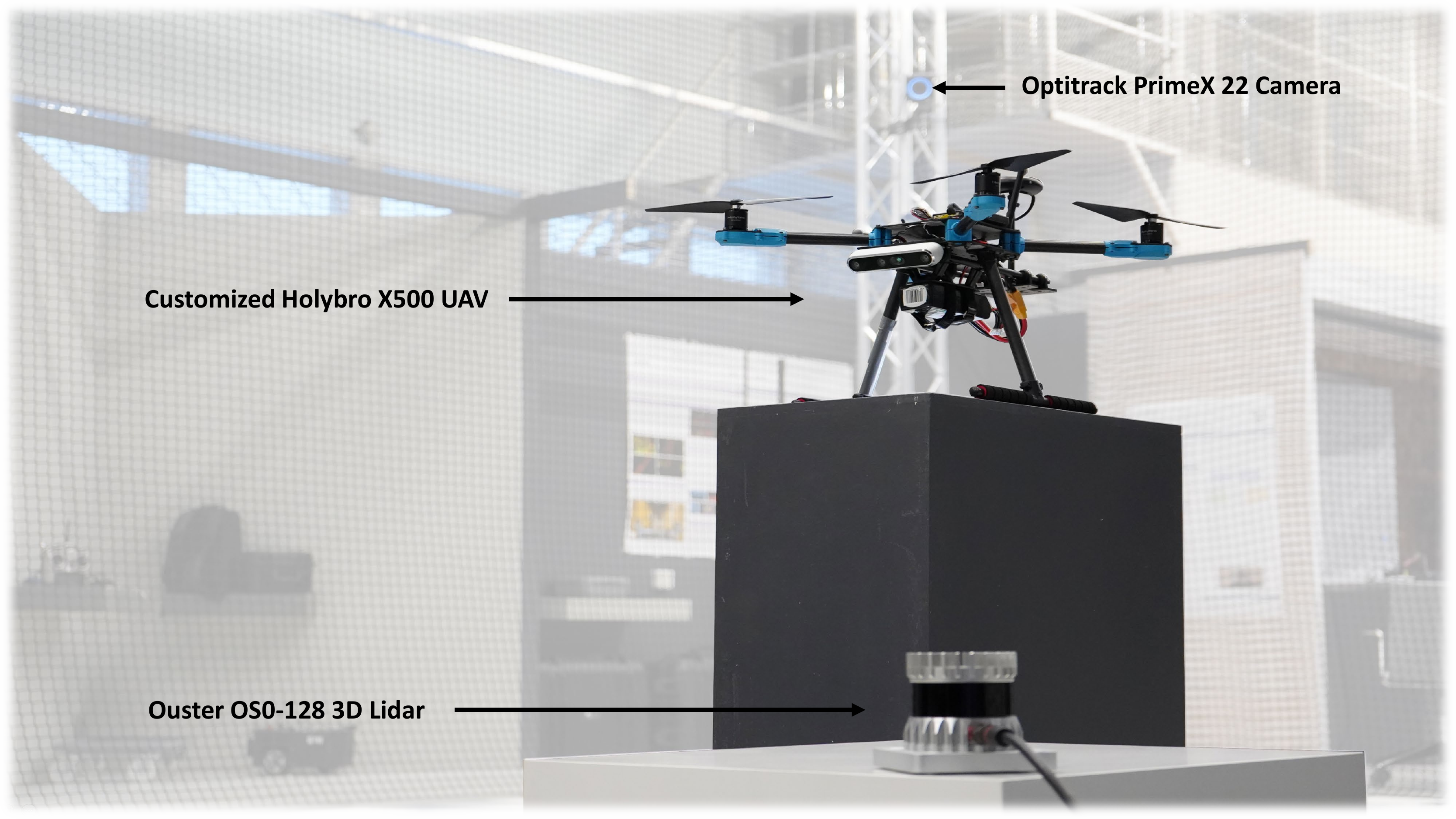}  
    \caption{Experimental hardware and site.}
    \label{fig:track_device} 
\end{figure}

The experimental setup consists of an Ouster OS0-128 LiDAR, an Intel computer, and a Holybro X500 V2 drone  shown in Fig.~\ref{fig:track_device}. The Ouster OS0-128 boasts a wide field of view ($360^\circ \times 90^\circ$) and is capable of producing both dense point cloud data and signal images at a frequency of 10\,Hz. The drone is equipped with OptiTrack markers, allowing the acquisition of the drone's actual position at 100\,Hz in the motion capture (MOCAP) system, which is also partially visible in Fig.~\ref{fig:track_device}.

% The Intel computer has 16\,GB RAM, a 6-core Intel i5-9300H (2.40\,GHz) processor, and an Nvidia GTX 1660Ti (1536 CUDA cores, 6\,GB VRAM).

\subsection{Software Information}

The system was implemented based on the ROS Noetic framework on Ubuntu 20.04 operating system. The tracking package, Ouster drivers, and OptiTrack mocap program were executed on the laptop computer connected to the Ouster LiDAR. Our tracking approach requires YOLOv5~\footnote{\href{https://github.com/ultralytics/yolov5/releases}{https://github.com/ultralytics/yolov5/releases}} for UAV detection and Open3D~\footnote{\href{http://www.open3d.org/}{http://www.open3d.org/}} for point cloud data processing. The algorithms and code designed and developed for these experiments are written in Python and are publicly available in a GitHub repository~\footnote{\href{http://github.com}{https://github.com/TIERS/UAV-tracking-based-on-LiDAR-as-a-camera}}.
% he computer ran the OptiTrack receiver node program, the Ouster LiDAR driver, and our MAV tracking package. 

% The Ouster LiDAR driver provides real-time, 1:1 spatially synchronized point cloud data and signal images, ensuring that every pixel in the 2D structured data corresponds to a 3D point in the LiDAR data without any discretization or resampling occurring. 
% The latter was implemented as a ROS node and capable of simultaneously processing point cloud data and signal images in real time. 
% The position of the UAV was extracted from the point cloud and laser radar signal image using a point cloud library (Open3D) and image object detection (YOLOV5), respectively.

\subsection{Data collection}\label{subsec:data}

Regarding the data consisting of UAV detections with the Ouster LiDAR, we collected three different data sequences ($Seq \quad i, i \in (1 \sim 3)$).
% \textit{Seq 1}, \textit{Seq 2}, and \textit{Seq 3}) 
in an indoor area of $10.0 \times 10.0 m^2$, with distances ranging 0.5\,m to 8\,m between the LiDAR and the UAV. The details of the collected data can be seen in Table~\ref{tab:sequence_detail}. \textit{Seq\,1} and \textit{Seq\,3} represent a helical ascension trajectory, while \textit{Seq 2} represents an elliptical trajectory. 
 
\begin{table}[htb]
\centering
\caption{Data sequences collected in our experiment. }
\resizebox{\linewidth}{!}{%
\scriptsize{
\begin{tabular}{@{}lcccc@{}}
\toprule
\textbf{Sequences} & \textbf{Time (s)} & \textbf{Ground Truth} &\textbf{Trajectory} & \textbf{Distance (m)} \\ \hline
\textit{Seq 1}& 35.8 & Mocap  &  elliptical trajectory & 7.0    \\
\textit{Seq 2} & 26.9 &Mocap  & spiral trajectory & 6.3   \\
\textit{Seq 3} &32.7 & Mocap  & spiral trajectory & 8.0   \\
\bottomrule
\end{tabular}
}
}
    \label{tab:sequence_detail}   
\end{table}

\subsection{UAV tracking fusing signal images and point clouds}

This manuscript introduces an approach to tracking a UAV by fusing LiDAR signal images and point cloud data, both generated within the same sensor. The proposed methodology's overarching framework is illustrated in Fig.~\ref{fig:concept}. It is worth noting that the Ouster OS0-128 LiDAR serves as the sole input source for the entire system, providing real-time UAV position outputs. The tracking procedure comprises two primary stages, namely:

% (i) Obtain the initial position of the UAV
\subsubsection{Initialization of UAV position}

\begin{figure}[b] 
    \centering   
    \includegraphics[width=0.45\textwidth]{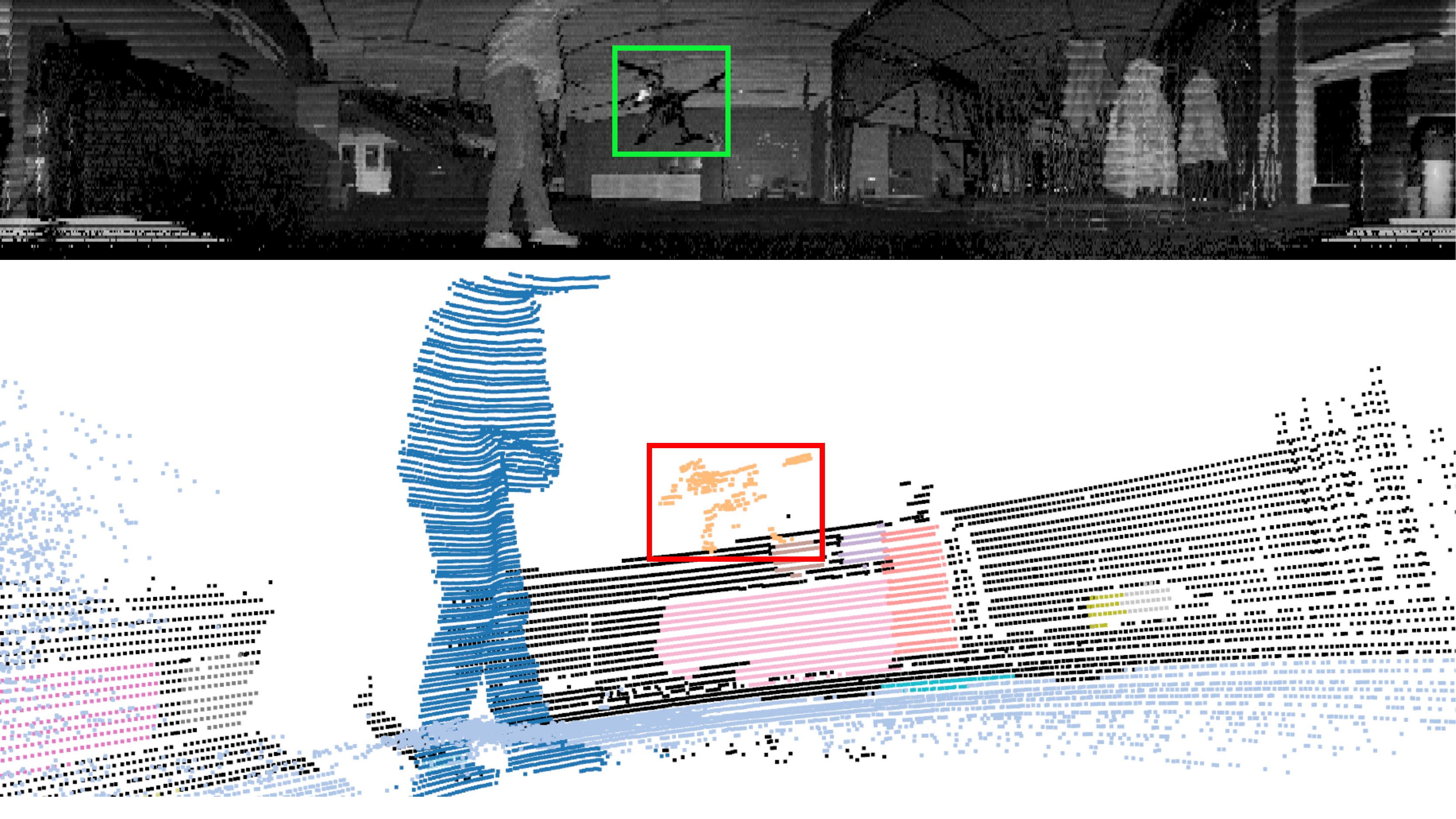}  
    \caption{Example of a signal image (top) and its corresponding point cloud with background removed (bottom).}
    \label{fig:signal_image} 
\end{figure}

Fig~\ref{fig:signal_image} shows the signal image and point cloud data of a UAV at its initial position. Notably, when the UAV approaches the Ouster LiDAR, the signal image of the UAV appears clearer. To detect the position of the UAV in the signal image and obtain the ROI in the image, the state-of-the-art object detection algorithm YOLOV5 is utilized. Given that the Ouster LiDAR signal image and the point cloud data are spatially linked, the corresponding point cloud ROI can be extracted. Subsequently, by employing ground removal and point cloud clustering techniques, the UAV point cloud can be extracted and, as such, the initial position of the UAV can be estimated.

% (ii) fusing signal images and point clouds
\subsubsection{Fusion of signal image and point cloud data}

Our goal for fusing the signal image and point cloud data is to acquire an accurate ROI. Initially, we perform image detection on each signal image to identify the UAV. This allows us to obtain a more precise ROI. Additionally, we can use the UAV's initial position from the previous step as a reference to extract the UAV point cloud from the environment based on the number of UAV point clouds and their distance. However, if YOLOv5 fails to detect the UAV, we select the ROI predicted by the Kalman filter and separate the UAV point cloud from it. This process is explicated in Algorithm~\ref{alg:fusing_data}.

\begin{algorithm}[t]
    \footnotesize
	\caption{UAV tracking fusing signal images and point clouds} 
	\label{alg:fusing_data}
	\KwIn{ \\
	    \vspace{.42em}
	    \hspace{.5em}Raw pointcloud:  $\mathcal{P}_{raw}^t$ \\
                  
	    \vspace{.23em}
	    \hspace{.5em}Signal image:  $\mathcal{S}^{t} $ \\

            \vspace{.23em}
	    \hspace{.5em}Target UAV point cloud:  $\mathcal{P}_{UAV}^{t-1}$ \\

	}
	\KwOut{  \\
	    \vspace{.42em}
	    \hspace{.5em}Drone pose: $\textbf{P}_{UAV}^{t}$\;
	    % \vspace{.23em}
        }
    \SetKwFunction{FMain}{object\_extraction}
    \SetKwProg{Fn}{Function}{:}{}
    \Fn{\FMain{$\mathcal{P}_{raw}^t$, $\mathcal{P}_{UAV}^{t-1}$, $\mathcal{ROI}^t_{YOLO}$}}{
        \If{$\mathcal{ROI}^t_{YOLO}$}{
            $\mathcal{P}_{roi}^t$ = $\mathcal{P}_{raw}^t$ ($\mathcal{ROI}^t_{YOLO}$)\;
        }
        \Else{
            $\mathcal{ROI}^t \longleftarrow KF \left( get\_center\left( \mathcal{P}_{UAV}^{t-1} \right) \right)$\;
            
            % $\mathcal{P}_{roi}^t$ \longleftarrow\ $\mathcal{P}^t_{raw}$ ( $\mathcal{ROI}^{t}$)\;
        }
        $\mathcal{P}^t \longleftarrow ground\_removal\left( \mathcal{P}_{ROI}^t \right)$ \;
        $\mathcal{P}_{i}^t \longleftarrow DBSCAN \left( \mathcal{P}^t \right),\:i\in(0,R)$\;
        \ForEach{ $\mathcal{P}$ $\in$ $\mathcal{P}_{i}^t$ }{%
            \If{ Min (num($\mathcal{P}$ ) - num($\mathcal{P}_{UAV}^{t-1}$) )}{%
                \If{Min (dis($\mathcal{P}$) - dis($\mathcal{P}_{UAV}^{t-1}$))}{%
                    $flag = 1$ \;
                    $\mathcal{P}^{t}_{uav} \longleftarrow \mathcal{P}$ \;
                }
                
            \Else{
                $flag = 0$\;
                }
            }
            \Else{ 
                $flag = 0$\;}
        }
    \KwRet $\mathcal{P}_{uav}^{t}$, flag \;
    }
    \ForEach{ \textit{new} $\mathcal{S}^t$ }{
        $\mathcal{ROI}^t_{YOLO} \longleftarrow\ YOLOv5 \left( \mathcal{S}^{t} \right)$\;
        \If{$\mathcal{ROI}^t_{YOLO} = \textbf{None}$}{
            $\mathcal{P}_{UAV}^{t}$ , flag = \textit{object\_extraction} ($\mathcal{P}_{raw}^t$, $\mathcal{P}_{uav}^{t-1}$) \;
            \If{flag = 0}{
                $\textbf{P}_{UAV}^{t}$ = \textit{KF\_predict} (\textit{get\_center}($\mathcal{P}_{UAV}^{t-1}$))  \;
                \textit{KF\_update} ($\textbf{P}_{UAV}^{t}$)
            }
            \Else{
                $\textbf{P}_{UAV}^{t}$ = \textit{get\_center}($\mathcal{P}_{UAV}^{t}$)  \;
                \textit{KF\_update} ($\textbf{P}_{UAV}^{t}$)  \;
            }
        }
        \Else{
            $\mathcal{P}_{UAV}^{t} = object\_extraction \left( \mathcal{P}_{raw}^t, \:\mathcal{P}_{UAV}^{t-1}, \:\mathcal{ROI}^t \right)$ \;
            $\textbf{P}_{UAV}^{t} = get\_center \left(\mathcal{P}_{UAV}^{t} \right)$\;
            $KF\_update \left( \textbf{P}_{UAV}^{t} \right)$\;
        }
    }
\end{algorithm}

\subsection{Evaluation}
% In order to validate the accuracy of the estimated poses and velocities of the UAV, we calculated the absolute pose error (APE) and velocity error based on the ground truth from the mocap system. Additionally, as Jetson Nano is a popular mobile computing platform, we  performed our tracking program in it to evaluate the computation consumption as a reference.
To validate the precision of the UAV estimated poses and velocities by our approach, we calculate the absolute pose error (APE) and velocity error based on the ground truth from the MOCAP system. We conducted a comparative analysis of our proposed method with a UAV tracking method that solely relies on either Ouster LiDAR images or point clouds.
The point cloud tracking method uses only Ouster OS0-128 LiDAR point cloud data as input, with a frame rate of\,10Hz. When tracking the UAV using point cloud data, the initial position of the UAV needs to be known as the point cloud of the UAV is sparser than that of larger objects, such as cars or humans, and distinguishing the point cloud of the UAV from the environment using features is challenging. On the other hand, the image tracking method uses only Ouster OS0-128 LiDAR signal images with a frame rate of 10\,Hz. Firstly, the signal image undergoes target detection processing to obtain the UAV's bounding box in the signal image. Subsequently, the image in the bounding box is converted into point cloud data. The point cloud clustering algorithm is then utilized to separate the UAV's point cloud from the environment based on the number and distance features of the point cloud clusters. This approach allows us to obtain the trajectory of the UAV. Both of these methods are estimated by the Kalman filter method to obtain the UAV's trajectory.

We conducted the experiments on two different platforms to assess real-time performance, the Lenovo Legion Y7000P equipped with 16GB RAM, 6-core Intel i5-9300H (2.40GHz) and Nvidia GTX 1660Ti (1536 CUDA cores, 6GB VRAM), as well as the commonly used embedded computing platform Jetson Nano with 4-core ARM A57 64-bit CPU (1.43GHz), 4GB RAM, and 128-core Maxwell GPU. 

% Furthermore, in addition to the Intel laptop, we also implemented our tracking program on the Jetson Nano, a widely used mobile computing platform, to assess its computational efficiency. 
%The Jetson Nano has 4G RAM and Quad-core ARM Cortex-A57 64-bit with 1.43\,GHz. The evaluations were operated with all three collected data sequences described in~\ref{subsec:data}.

% \newpage
%%%%%%%%%%%%%%%%%%%%%%%%%%%%%%%%%%%%%%%%%%%%%%
%%                                          %%
%%              EXPERIMENTS                 %%
%%                                          %%
%%%%%%%%%%%%%%%%%%%%%%%%%%%%%%%%%%%%%%%%%%%%%%

\begin{figure}[ht]
    \centering
    \includegraphics[width=0.495\textwidth]{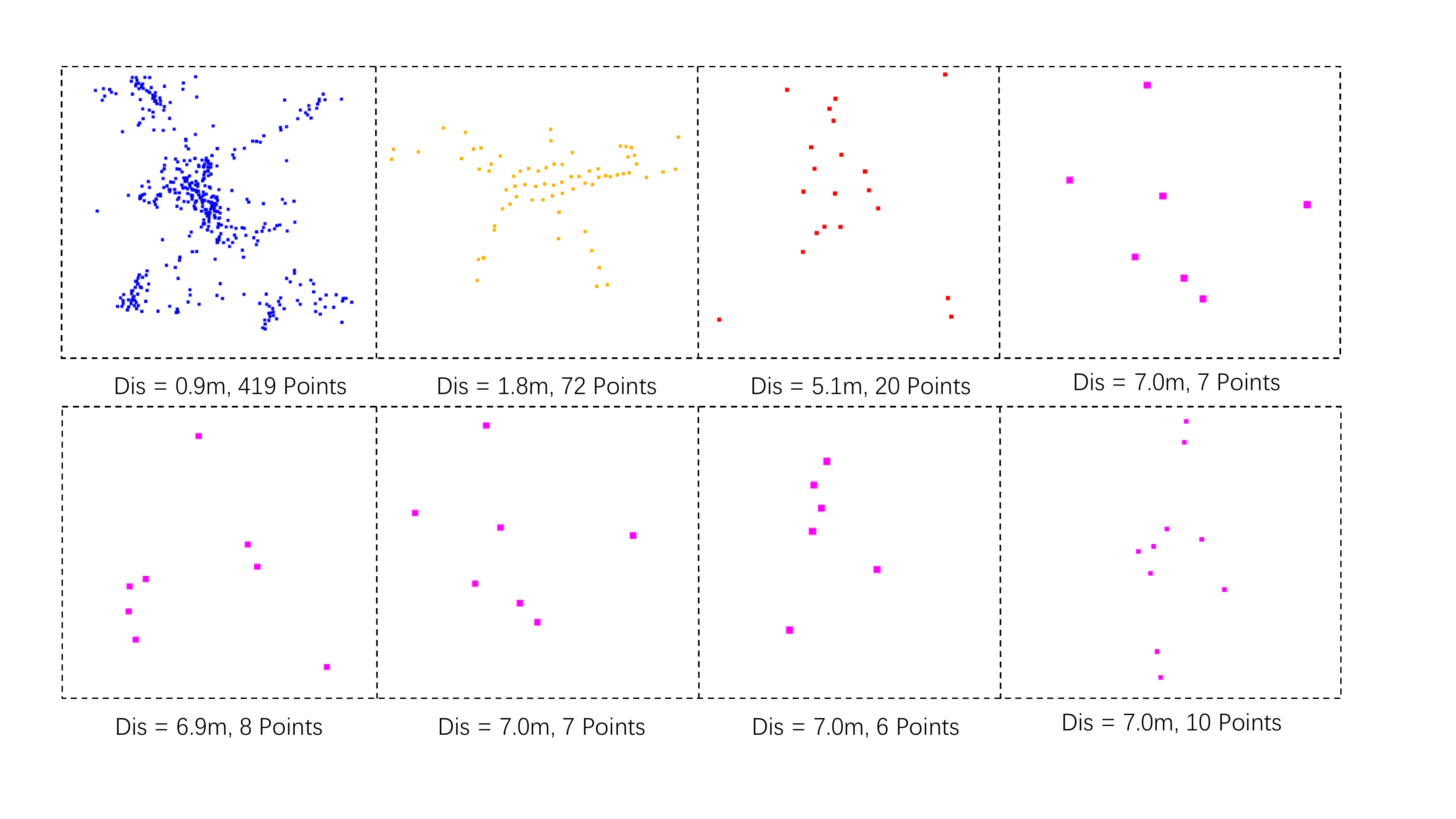}
    % \caption{UAV in point cloud}
    \caption{UAV point cloud of drones at different distances, the bottom line shows UAV point cloud of four consecutive frames at the same long distance.}
    \label{fig:pcd_frame}
\end{figure}

\begin{figure*}[ht]
    \begin{subfigure}{.32\textwidth}
        \centering
        \setlength{\figurewidth}{\textwidth}
        \setlength{\figureheight}{1.05\textwidth}
        \scriptsize{% This file was created with tikzplotlib v0.9.14.
\begin{tikzpicture}

% \definecolor{color0}{rgb}{0.6,0,0.980392156862745}
% \definecolor{color1}{rgb}{0.968627450980392,0.00784313725490196,0.164705882352941}
% \definecolor{color2}{rgb}{0.0235294117647059,0.32156862745098,1}

\definecolor{color0}{rgb}{0.90, 0.62, 0.00}
\definecolor{color1}{rgb}{0.34, 0.70, 0.91}
\definecolor{color2}{rgb}{0.00, 0.62, 0.45}
\definecolor{color3}{rgb}{0.94, 0.89, 0.27}
\definecolor{color4}{rgb}{0.00, 0.45, 0.69}
\definecolor{color5}{rgb}{0.83, 0.37, 0.00}

\begin{axis}[
    width=\figurewidth,
    height=\figureheight,
    legend cell align={left},
    legend style={
      fill opacity=0.8,
      draw opacity=1,
      text opacity=1,
      at={(-0.03,1.10)},
      anchor=north west,
      draw=white,
      /tikz/every even column/.append style={column sep=0.5cm},
    },
    legend columns=3,
    axis line style={white},
    tick align=outside,
    tick pos=left,
    x grid style={white!69.0196078431373!black},
    xmin=0.5, xmax=6.3,
    xtick style={color=black},
    xtick={1.4,3.4,5.4},
    xticklabels={X,Y,Z},
    y grid style={white!69.0196078431373!black},
    ylabel={ Error (m)},
    ymin=-0.0147050240191961, ymax=0.31,
    ytick style={color=black},
    xmajorgrids,
    % xmajorticks=true,
    % xmin=-0.5, xmax=3.3125,
    xminorgrids,
    ymajorgrids,
    ymajorticks=true,
    minor y tick num = 3,
    minor y grid style={dashed},
    yminorgrids,
    yticklabel style={
            /pgf/number format/fixed,
            /pgf/number format/precision=5
    },
    scaled y ticks=false,
]

\draw[draw=color0,fill=color0, thick] (axis cs:0,0) rectangle (axis cs:0,0);
\addlegendimage{ybar, ybar legend, draw=color0,fill=color0,thick}
\addlegendentry{Tracking with only point cloud}

\draw[draw=color1,fill=color1, thick] (axis cs:0,0) rectangle (axis cs:0,0);
\addlegendimage{ybar, ybar legend, draw=color1,fill=color1,thick}
\addlegendentry{Tracking with only signal images}

\draw[draw=color2,fill=color2, thick] (axis cs:0,0) rectangle (axis cs:0,0);
\addlegendimage{ybar, ybar legend, draw=color2,fill=color2,thick}
\addlegendentry{Our approach}

\path [draw=color0, fill=color0]
(axis cs:0.9,0.0359016596706485)
--(axis cs:1.1,0.0359016596706485)
--(axis cs:1.1,0.123964853476531)
--(axis cs:0.9,0.123964853476531)
--(axis cs:0.9,0.0359016596706485)
--cycle;
\addplot [black, forget plot]
table {%
1 0.0359016596706485
1 0.000594158118181198
};
\addplot [black, forget plot]
table {%
1 0.123964853476531
1 0.219165955456631
};
\addplot [black, forget plot]
table {%
0.95 0.000594158118181198
1.05 0.000594158118181198
};
\addplot [black, forget plot]
table {%
0.95 0.219165955456631
1.05 0.219165955456631
};
\path [draw=color1, fill=color1]
(axis cs:1.3,0.0224564700129198)
--(axis cs:1.5,0.0224564700129198)
--(axis cs:1.5,0.131575909470591)
--(axis cs:1.3,0.131575909470591)
--(axis cs:1.3,0.0224564700129198)
--cycle;
\addplot [black, forget plot]
table {%
1.4 0.0224564700129198
1.4 0.00368735991092217
};
\addplot [black, forget plot]
table {%
1.4 0.131575909470591
1.4 0.177035822042366
};
\addplot [black, forget plot]
table {%
1.35 0.00368735991092217
1.45 0.00368735991092217
};
\addplot [black, forget plot]
table {%
1.35 0.177035822042366
1.45 0.177035822042366
};
\path [draw=color2, fill=color2]
(axis cs:1.7,0.00882856225400375)
--(axis cs:1.9,0.00882856225400375)
--(axis cs:1.9,0.0287960855190496)
--(axis cs:1.7,0.0287960855190496)
--(axis cs:1.7,0.00882856225400375)
--cycle;
\addplot [black, forget plot]
table {%
1.8 0.00882856225400375
1.8 0.000267686459507033
};
\addplot [black, forget plot]
table {%
1.8 0.0287960855190496
1.8 0.0584475453119366
};
\addplot [black, forget plot]
table {%
1.75 0.000267686459507033
1.85 0.000267686459507033
};
\addplot [black, forget plot]
table {%
1.75 0.0584475453119366
1.85 0.0584475453119366
};
\path [draw=color0, fill=color0]
(axis cs:2.9,0.0365788961205427)
--(axis cs:3.1,0.0365788961205427)
--(axis cs:3.1,0.0941762357336027)
--(axis cs:2.9,0.0941762357336027)
--(axis cs:2.9,0.0365788961205427)
--cycle;
\addplot [black, forget plot]
table {%
3 0.0365788961205427
3 0.000869162330838336
};
\addplot [black, forget plot]
table {%
3 0.0941762357336027
3 0.17826472756505
};
\addplot [black, forget plot]
table {%
2.95 0.000869162330838336
3.05 0.000869162330838336
};
\addplot [black, forget plot]
table {%
2.95 0.17826472756505
3.05 0.17826472756505
};
\path [draw=color1, fill=color1]
(axis cs:3.3,0.0936335797918026)
--(axis cs:3.5,0.0936335797918026)
--(axis cs:3.5,0.225292738533558)
--(axis cs:3.3,0.225292738533558)
--(axis cs:3.3,0.0936335797918026)
--cycle;
\addplot [black, forget plot]
table {%
3.4 0.0936335797918026
3.4 0.00413021189512097
};
\addplot [black, forget plot]
table {%
3.4 0.225292738533558
3.4 0.294725109117304
};
\addplot [black, forget plot]
table {%
3.35 0.00413021189512097
3.45 0.00413021189512097
};
\addplot [black, forget plot]
table {%
3.35 0.294725109117304
3.45 0.294725109117304
};
\path [draw=color2, fill=color2]
(axis cs:3.7,0.0105550124701859)
--(axis cs:3.9,0.0105550124701859)
--(axis cs:3.9,0.0439037283093395)
--(axis cs:3.7,0.0439037283093395)
--(axis cs:3.7,0.0105550124701859)
--cycle;
\addplot [black, forget plot]
table {%
3.8 0.0105550124701859
3.8 0.000138631389560828
};
\addplot [black, forget plot]
table {%
3.8 0.0439037283093395
3.8 0.0823405821542753
};
\addplot [black, forget plot]
table {%
3.75 0.000138631389560828
3.85 0.000138631389560828
};
\addplot [black, forget plot]
table {%
3.75 0.0823405821542753
3.85 0.0823405821542753
};
\path [draw=color0, fill=color0]
(axis cs:4.9,0.0255128619703696)
--(axis cs:5.1,0.0255128619703696)
--(axis cs:5.1,0.0756290965564553)
--(axis cs:4.9,0.0756290965564553)
--(axis cs:4.9,0.0255128619703696)
--cycle;
\addplot [black, forget plot]
table {%
5 0.0255128619703696
5 0.00148441613567574
};
\addplot [black, forget plot]
table {%
5 0.0756290965564553
5 0.148283764805501
};
\addplot [black, forget plot]
table {%
4.95 0.00148441613567574
5.05 0.00148441613567574
};
\addplot [black, forget plot]
table {%
4.95 0.148283764805501
5.05 0.148283764805501
};
\path [draw=color1, fill=color1]
(axis cs:5.3,0.00532853403802855)
--(axis cs:5.5,0.00532853403802855)
--(axis cs:5.5,0.0127842448158017)
--(axis cs:5.3,0.0127842448158017)
--(axis cs:5.3,0.00532853403802855)
--cycle;
\addplot [black, forget plot]
table {%
5.4 0.00532853403802855
5.4 0.00142377354419265
};
\addplot [black, forget plot]
table {%
5.4 0.0127842448158017
5.4 0.0237795656365393
};
\addplot [black, forget plot]
table {%
5.35 0.00142377354419265
5.45 0.00142377354419265
};
\addplot [black, forget plot]
table {%
5.35 0.0237795656365393
5.45 0.0237795656365393
};
\path [draw=color2, fill=color2]
(axis cs:5.7,0.00727972918056854)
--(axis cs:5.9,0.00727972918056854)
--(axis cs:5.9,0.0231674857426256)
--(axis cs:5.7,0.0231674857426256)
--(axis cs:5.7,0.00727972918056854)
--cycle;
\addplot [black, forget plot]
table {%
5.8 0.00727972918056854
5.8 2.9744225399142e-05
};
\addplot [black, forget plot]
table {%
5.8 0.0231674857426256
5.8 0.0459741988826368
};
\addplot [black, forget plot]
table {%
5.75 2.9744225399142e-05
5.85 2.9744225399142e-05
};
\addplot [black, forget plot]
table {%
5.75 0.0459741988826368
5.85 0.0459741988826368
};
\addplot [line width=1pt, white, forget plot]
table {%
0.9 0.0765784603353681
1.1 0.0765784603353681
};
\addplot [line width=1pt, white, forget plot]
table {%
1.3 0.0798680383669237
1.5 0.0798680383669237
};
\addplot [line width=1pt, white, forget plot]
table {%
1.7 0.0167689702781684
1.9 0.0167689702781684
};
\addplot [line width=1pt, white, forget plot]
table {%
2.9 0.0651368712573408
3.1 0.0651368712573408
};
\addplot [line width=1pt, white, forget plot]
table {%
3.3 0.167468870553379
3.5 0.167468870553379
};
\addplot [line width=1pt, white, forget plot]
table {%
3.7 0.0247040115872634
3.9 0.0247040115872634
};
\addplot [line width=1pt, white, forget plot]
table {%
4.9 0.0489386840818955
5.1 0.0489386840818955
};
\addplot [line width=1pt, white, forget plot]
table {%
5.3 0.00967427726703923
5.5 0.00967427726703923
};
\addplot [line width=1pt, white, forget plot]
table {%
5.7 0.0139552293806988
5.9 0.0139552293806988
};
\end{axis}

\end{tikzpicture}}
        \caption{\scriptsize{Seq\,1}}
        \label{fig:err_seq1}
    \end{subfigure}
    \hfill
    \begin{subfigure}{.32\textwidth}
        \centering
        \setlength{\figurewidth}{\textwidth}
        \setlength{\figureheight}{1.03\textwidth}
        \scriptsize{% This file was created with tikzplotlib v0.9.14.
\begin{tikzpicture}

% \definecolor{color0}{rgb}{0.6,0,0.980392156862745}
% \definecolor{color1}{rgb}{0.968627450980392,0.00784313725490196,0.164705882352941}
% \definecolor{color2}{rgb}{0.0235294117647059,0.32156862745098,1}
\definecolor{color0}{rgb}{0.90, 0.62, 0.00}
\definecolor{color1}{rgb}{0.34, 0.70, 0.91}
\definecolor{color2}{rgb}{0.00, 0.62, 0.45}
\definecolor{color3}{rgb}{0.94, 0.89, 0.27}
\definecolor{color4}{rgb}{0.00, 0.45, 0.69}
\definecolor{color5}{rgb}{0.83, 0.37, 0.00}

\begin{axis}[
    width=\figurewidth,
    height=\figureheight,
    legend cell align={left},
    legend style={
      fill opacity=0.8,
      draw opacity=1,
      text opacity=1,
      at={(0.03,0.97)},
      anchor=north west,
      draw=white!80!black
    },
axis line style={white},
tick align=outside,
tick pos=left,
x grid style={white!69.0196078431373!black},
xmin=0.5, xmax=6.3,
xtick style={color=black},
xtick={1.4,3.4,5.4},
xticklabels={X,Y,Z},
y grid style={white!69.0196078431373!black},
ylabel={ Error (m)},
ymin=-0.00876739680516033, ymax=0.184969536254822,
ytick style={color=black},
    xmajorgrids,
    % xmajorticks=true,
    % xmin=-0.5, xmax=3.3125,
    xminorgrids,
    ymajorgrids,
    ymajorticks=true,
    minor y tick num = 3,
    minor y grid style={dashed},
    yminorgrids,
    yticklabel style={
            /pgf/number format/fixed,
            /pgf/number format/precision=5
    },
    scaled y ticks=false,
]
\path [draw=color0, fill=color0]
(axis cs:0.9,0.00815245852515445)
--(axis cs:1.1,0.00815245852515445)
--(axis cs:1.1,0.0355660872611989)
--(axis cs:0.9,0.0355660872611989)
--(axis cs:0.9,0.00815245852515445)
--cycle;
\addplot [black, forget plot]
table {%
1 0.00815245852515445
1 0.000677464621831714
};
\addplot [black, forget plot]
table {%
1 0.0355660872611989
1 0.0760465676788473
};
\addplot [black, forget plot]
table {%
0.95 0.000677464621831714
1.05 0.000677464621831714
};
\addplot [black, forget plot]
table {%
0.95 0.0760465676788473
1.05 0.0760465676788473
};
\path [draw=color1, fill=color1]
(axis cs:1.3,0.00430830867868714)
--(axis cs:1.5,0.00430830867868714)
--(axis cs:1.5,0.0151839841291762)
--(axis cs:1.3,0.0151839841291762)
--(axis cs:1.3,0.00430830867868714)
--cycle;
\addplot [black, forget plot]
table {%
1.4 0.00430830867868714
1.4 0.0010557370823836
};
\addplot [black, forget plot]
table {%
1.4 0.0151839841291762
1.4 0.0231649600846127
};
\addplot [black, forget plot]
table {%
1.35 0.0010557370823836
1.45 0.0010557370823836
};
\addplot [black, forget plot]
table {%
1.35 0.0231649600846127
1.45 0.0231649600846127
};
\path [draw=color2, fill=color2]
(axis cs:1.7,0.0108070949987866)
--(axis cs:1.9,0.0108070949987866)
--(axis cs:1.9,0.0469455026016019)
--(axis cs:1.7,0.0469455026016019)
--(axis cs:1.7,0.0108070949987866)
--cycle;
\addplot [black, forget plot]
table {%
1.8 0.0108070949987866
1.8 3.88274248388498e-05
};
\addplot [black, forget plot]
table {%
1.8 0.0469455026016019
1.8 0.0722655397008891
};
\addplot [black, forget plot]
table {%
1.75 3.88274248388498e-05
1.85 3.88274248388498e-05
};
\addplot [black, forget plot]
table {%
1.75 0.0722655397008891
1.85 0.0722655397008891
};
\path [draw=color0, fill=color0]
(axis cs:2.9,0.00974226341835482)
--(axis cs:3.1,0.00974226341835482)
--(axis cs:3.1,0.0424458183453618)
--(axis cs:2.9,0.0424458183453618)
--(axis cs:2.9,0.00974226341835482)
--cycle;
\addplot [black, forget plot]
table {%
3 0.00974226341835482
3 0.0003648774884768
};
\addplot [black, forget plot]
table {%
3 0.0424458183453618
3 0.0759341926763533
};
\addplot [black, forget plot]
table {%
2.95 0.0003648774884768
3.05 0.0003648774884768
};
\addplot [black, forget plot]
table {%
2.95 0.0759341926763533
3.05 0.0759341926763533
};
\path [draw=color1, fill=color1]
(axis cs:3.3,0.0336190006785358)
--(axis cs:3.5,0.0336190006785358)
--(axis cs:3.5,0.097577699413025)
--(axis cs:3.3,0.097577699413025)
--(axis cs:3.3,0.0336190006785358)
--cycle;
\addplot [black, forget plot]
table {%
3.4 0.0336190006785358
3.4 0.0030644107688782
};
\addplot [black, forget plot]
table {%
3.4 0.097577699413025
3.4 0.176163312024822
};
\addplot [black, forget plot]
table {%
3.35 0.0030644107688782
3.45 0.0030644107688782
};
\addplot [black, forget plot]
table {%
3.35 0.176163312024822
3.45 0.176163312024822
};
\path [draw=color2, fill=color2]
(axis cs:3.7,0.00808186866347871)
--(axis cs:3.9,0.00808186866347871)
--(axis cs:3.9,0.0440748566983342)
--(axis cs:3.7,0.0440748566983342)
--(axis cs:3.7,0.00808186866347871)
--cycle;
\addplot [black, forget plot]
table {%
3.8 0.00808186866347871
3.8 0.0001008556550679
};
\addplot [black, forget plot]
table {%
3.8 0.0440748566983342
3.8 0.0881248202830607
};
\addplot [black, forget plot]
table {%
3.75 0.0001008556550679
3.85 0.0001008556550679
};
\addplot [black, forget plot]
table {%
3.75 0.0881248202830607
3.85 0.0881248202830607
};
\path [draw=color0, fill=color0]
(axis cs:4.9,0.00612243655419784)
--(axis cs:5.1,0.00612243655419784)
--(axis cs:5.1,0.0209237401912207)
--(axis cs:4.9,0.0209237401912207)
--(axis cs:4.9,0.00612243655419784)
--cycle;
\addplot [black, forget plot]
table {%
5 0.00612243655419784
5 0.000388261695153491
};
\addplot [black, forget plot]
table {%
5 0.0209237401912207
5 0.0392731995430584
};
\addplot [black, forget plot]
table {%
4.95 0.000388261695153491
5.05 0.000388261695153491
};
\addplot [black, forget plot]
table {%
4.95 0.0392731995430584
5.05 0.0392731995430584
};
\path [draw=color1, fill=color1]
(axis cs:5.3,0.0139338719068252)
--(axis cs:5.5,0.0139338719068252)
--(axis cs:5.5,0.040402023571507)
--(axis cs:5.3,0.040402023571507)
--(axis cs:5.3,0.0139338719068252)
--cycle;
\addplot [black, forget plot]
table {%
5.4 0.0139338719068252
5.4 0.00455167088083763
};
\addplot [black, forget plot]
table {%
5.4 0.040402023571507
5.4 0.065507495128893
};
\addplot [black, forget plot]
table {%
5.35 0.00455167088083763
5.45 0.00455167088083763
};
\addplot [black, forget plot]
table {%
5.35 0.065507495128893
5.45 0.065507495128893
};
\path [draw=color2, fill=color2]
(axis cs:5.7,0.004761184148176)
--(axis cs:5.9,0.004761184148176)
--(axis cs:5.9,0.0178303404227792)
--(axis cs:5.7,0.0178303404227792)
--(axis cs:5.7,0.004761184148176)
--cycle;
\addplot [black, forget plot]
table {%
5.8 0.004761184148176
5.8 0.000167742242199065
};
\addplot [black, forget plot]
table {%
5.8 0.0178303404227792
5.8 0.0364272724509153
};
\addplot [black, forget plot]
table {%
5.75 0.000167742242199065
5.85 0.000167742242199065
};
\addplot [black, forget plot]
table {%
5.75 0.0364272724509153
5.85 0.0364272724509153
};
\addplot [line width=1pt, white, forget plot]
table {%
0.9 0.0202748343545978
1.1 0.0202748343545978
};
\addplot [line width=1pt, white, forget plot]
table {%
1.3 0.00982938124535071
1.5 0.00982938124535071
};
\addplot [line width=1pt, white, forget plot]
table {%
1.7 0.029147362000046
1.9 0.029147362000046
};
\addplot [line width=1pt, white, forget plot]
table {%
2.9 0.0267552841689922
3.1 0.0267552841689922
};
\addplot [line width=1pt, white, forget plot]
table {%
3.3 0.0701316623582899
3.5 0.0701316623582899
};
\addplot [line width=1pt, white, forget plot]
table {%
3.7 0.0192756859440788
3.9 0.0192756859440788
};
\addplot [line width=1pt, white, forget plot]
table {%
4.9 0.0129318118170228
5.1 0.0129318118170228
};
\addplot [line width=1pt, white, forget plot]
table {%
5.3 0.0231806449570045
5.5 0.0231806449570045
};
\addplot [line width=1pt, white, forget plot]
table {%
5.7 0.0113959344884098
5.9 0.0113959344884098
};
\end{axis}

\end{tikzpicture}}
        \caption{\scriptsize{Seq\,2}}
        \label{fig:err_seq2}     
    \end{subfigure}
    \hfill
    \begin{subfigure}{.32\textwidth}
        \centering
        \setlength{\figurewidth}{\textwidth}
        \setlength{\figureheight}{\textwidth}
        \scriptsize{% This file was created with tikzplotlib v0.9.14.
\begin{tikzpicture}

% \definecolor{color0}{rgb}{0.6,0,0.980392156862745}
% \definecolor{color1}{rgb}{0.968627450980392,0.00784313725490196,0.164705882352941}
% \definecolor{color2}{rgb}{0.0235294117647059,0.32156862745098,1}
\definecolor{color0}{rgb}{0.90, 0.62, 0.00}
\definecolor{color1}{rgb}{0.34, 0.70, 0.91}
\definecolor{color2}{rgb}{0.00, 0.62, 0.45}
\definecolor{color3}{rgb}{0.94, 0.89, 0.27}
\definecolor{color4}{rgb}{0.00, 0.45, 0.69}
\definecolor{color5}{rgb}{0.83, 0.37, 0.00}

\begin{axis}[
    width=\figurewidth,
    height=\figureheight,
    legend cell align={left},
    legend style={
      fill opacity=0.8,
      draw opacity=1,
      text opacity=1,
      at={(0.03,0.97)},
      anchor=north west,
      draw=white!80!black
    },
axis line style={white},
tick align=outside,
tick pos=left,
x grid style={white!69.0196078431373!black},
xmin=0.5, xmax=6.3,
xtick style={color=black},
xtick={1.4,3.4,5.4},
xticklabels={X,Y,Z},
y grid style={white!69.0196078431373!black},
ylabel={ Error (m)},
ymin=-0.00897775582982006, ymax=0.189937849485646,
ytick style={color=black},
        xmajorgrids,
    % xmajorticks=true,
    % xmin=-0.5, xmax=3.3125,
    xminorgrids,
    ymajorgrids,
    ymajorticks=true,
    minor y tick num = 3,
    minor y grid style={dashed},
    yminorgrids,
    yticklabel style={
            /pgf/number format/fixed,
            /pgf/number format/precision=5
    },
    scaled y ticks=false,
]
\path [draw=color0, fill=color0]
(axis cs:0.9,0.0232780218577666)
--(axis cs:1.1,0.0232780218577666)
--(axis cs:1.1,0.0711417617143559)
--(axis cs:0.9,0.0711417617143559)
--(axis cs:0.9,0.0232780218577666)
--cycle;
\addplot [black, forget plot]
table {%
1 0.0232780218577666
1 0.000596100612618056
};
\addplot [black, forget plot]
table {%
1 0.0711417617143559
1 0.142771459094593
};
\addplot [black, forget plot]
table {%
0.95 0.000596100612618056
1.05 0.000596100612618056
};
\addplot [black, forget plot]
table {%
0.95 0.142771459094593
1.05 0.142771459094593
};
\path [draw=color1, fill=color1]
(axis cs:1.3,0.0088396449768362)
--(axis cs:1.5,0.0088396449768362)
--(axis cs:1.5,0.0177479663700559)
--(axis cs:1.3,0.0177479663700559)
--(axis cs:1.3,0.0088396449768362)
--cycle;
\addplot [black, forget plot]
table {%
1.4 0.0088396449768362
1.4 0.000359157874370908
};
\addplot [black, forget plot]
table {%
1.4 0.0177479663700559
1.4 0.0278344120708334
};
\addplot [black, forget plot]
table {%
1.35 0.000359157874370908
1.45 0.000359157874370908
};
\addplot [black, forget plot]
table {%
1.35 0.0278344120708334
1.45 0.0278344120708334
};
\path [draw=color2, fill=color2]
(axis cs:1.7,0.0114199810676545)
--(axis cs:1.9,0.0114199810676545)
--(axis cs:1.9,0.053081363686204)
--(axis cs:1.7,0.053081363686204)
--(axis cs:1.7,0.0114199810676545)
--cycle;
\addplot [black, forget plot]
table {%
1.8 0.0114199810676545
1.8 0.000901600574375827
};
\addplot [black, forget plot]
table {%
1.8 0.053081363686204
1.8 0.108638698498432
};
\addplot [black, forget plot]
table {%
1.75 0.000901600574375827
1.85 0.000901600574375827
};
\addplot [black, forget plot]
table {%
1.75 0.108638698498432
1.85 0.108638698498432
};
\path [draw=color0, fill=color0]
(axis cs:2.9,0.0279910207296199)
--(axis cs:3.1,0.0279910207296199)
--(axis cs:3.1,0.0771662866628215)
--(axis cs:2.9,0.0771662866628215)
--(axis cs:2.9,0.0279910207296199)
--cycle;
\addplot [black, forget plot]
table {%
3 0.0279910207296199
3 0.000493392369484091
};
\addplot [black, forget plot]
table {%
3 0.0771662866628215
3 0.14312320811591
};
\addplot [black, forget plot]
table {%
2.95 0.000493392369484091
3.05 0.000493392369484091
};
\addplot [black, forget plot]
table {%
2.95 0.14312320811591
3.05 0.14312320811591
};
\path [draw=color1, fill=color1]
(axis cs:3.3,0.133592629973776)
--(axis cs:3.5,0.133592629973776)
--(axis cs:3.5,0.1742756192671)
--(axis cs:3.3,0.1742756192671)
--(axis cs:3.3,0.133592629973776)
--cycle;
\addplot [black, forget plot]
table {%
3.4 0.133592629973776
3.4 0.0787552872431321
};
\addplot [black, forget plot]
table {%
3.4 0.1742756192671
3.4 0.180896231062216
};
\addplot [black, forget plot]
table {%
3.35 0.0787552872431321
3.45 0.0787552872431321
};
\addplot [black, forget plot]
table {%
3.35 0.180896231062216
3.45 0.180896231062216
};
\path [draw=color2, fill=color2]
(axis cs:3.7,0.0143334506366791)
--(axis cs:3.9,0.0143334506366791)
--(axis cs:3.9,0.0484907984263832)
--(axis cs:3.7,0.0484907984263832)
--(axis cs:3.7,0.0143334506366791)
--cycle;
\addplot [black, forget plot]
table {%
3.8 0.0143334506366791
3.8 0.00152332401541067
};
\addplot [black, forget plot]
table {%
3.8 0.0484907984263832
3.8 0.0919856255534661
};
\addplot [black, forget plot]
table {%
3.75 0.00152332401541067
3.85 0.00152332401541067
};
\addplot [black, forget plot]
table {%
3.75 0.0919856255534661
3.85 0.0919856255534661
};
\path [draw=color0, fill=color0]
(axis cs:4.9,0.00718570042061395)
--(axis cs:5.1,0.00718570042061395)
--(axis cs:5.1,0.02510844997279)
--(axis cs:4.9,0.02510844997279)
--(axis cs:4.9,0.00718570042061395)
--cycle;
\addplot [black, forget plot]
table {%
5 0.00718570042061395
5 0.000149403716627639
};
\addplot [black, forget plot]
table {%
5 0.02510844997279
5 0.0515712221274507
};
\addplot [black, forget plot]
table {%
4.95 0.000149403716627639
5.05 0.000149403716627639
};
\addplot [black, forget plot]
table {%
4.95 0.0515712221274507
5.05 0.0515712221274507
};
\path [draw=color1, fill=color1]
(axis cs:5.3,0.016524061983732)
--(axis cs:5.5,0.016524061983732)
--(axis cs:5.5,0.0702343830365995)
--(axis cs:5.3,0.0702343830365995)
--(axis cs:5.3,0.016524061983732)
--cycle;
\addplot [black, forget plot]
table {%
5.4 0.016524061983732
5.4 0.00088228305353566
};
\addplot [black, forget plot]
table {%
5.4 0.0702343830365995
5.4 0.136660089747208
};
\addplot [black, forget plot]
table {%
5.35 0.00088228305353566
5.45 0.00088228305353566
};
\addplot [black, forget plot]
table {%
5.35 0.136660089747208
5.45 0.136660089747208
};
\path [draw=color2, fill=color2]
(axis cs:5.7,0.00499251809664381)
--(axis cs:5.9,0.00499251809664381)
--(axis cs:5.9,0.0189492854816761)
--(axis cs:5.7,0.0189492854816761)
--(axis cs:5.7,0.00499251809664381)
--cycle;
\addplot [black, forget plot]
table {%
5.8 0.00499251809664381
5.8 6.38625936102422e-05
};
\addplot [black, forget plot]
table {%
5.8 0.0189492854816761
5.8 0.0394391370956968
};
\addplot [black, forget plot]
table {%
5.75 6.38625936102422e-05
5.85 6.38625936102422e-05
};
\addplot [black, forget plot]
table {%
5.75 0.0394391370956968
5.85 0.0394391370956968
};
\addplot [line width=1pt, white, forget plot]
table {%
0.9 0.0452829561791654
1.1 0.0452829561791654
};
\addplot [line width=1pt, white, forget plot]
table {%
1.3 0.0120809718295072
1.5 0.0120809718295072
};
\addplot [line width=1pt, white, forget plot]
table {%
1.7 0.0255258806510643
1.9 0.0255258806510643
};
\addplot [line width=1pt, white, forget plot]
table {%
2.9 0.0515489334434402
3.1 0.0515489334434402
};
\addplot [line width=1pt, white, forget plot]
table {%
3.3 0.14480473897281
3.5 0.14480473897281
};
\addplot [line width=1pt, white, forget plot]
table {%
3.7 0.025519191420708
3.9 0.025519191420708
};
\addplot [line width=1pt, white, forget plot]
table {%
4.9 0.0157743244749928
5.1 0.0157743244749928
};
\addplot [line width=1pt, white, forget plot]
table {%
5.3 0.0370830731462402
5.5 0.0370830731462402
};
\addplot [line width=1pt, white, forget plot]
table {%
5.7 0.00942297550400267
5.9 0.00942297550400267
};
\end{axis}

\end{tikzpicture}}
        \caption{\scriptsize{Seq\,3}}
        \label{fig:err_seq3}     
    \end{subfigure}
    \caption{Absolute position error (APE) value of three data sequences.}
    \label{fig:ape_error} 
\end{figure*}
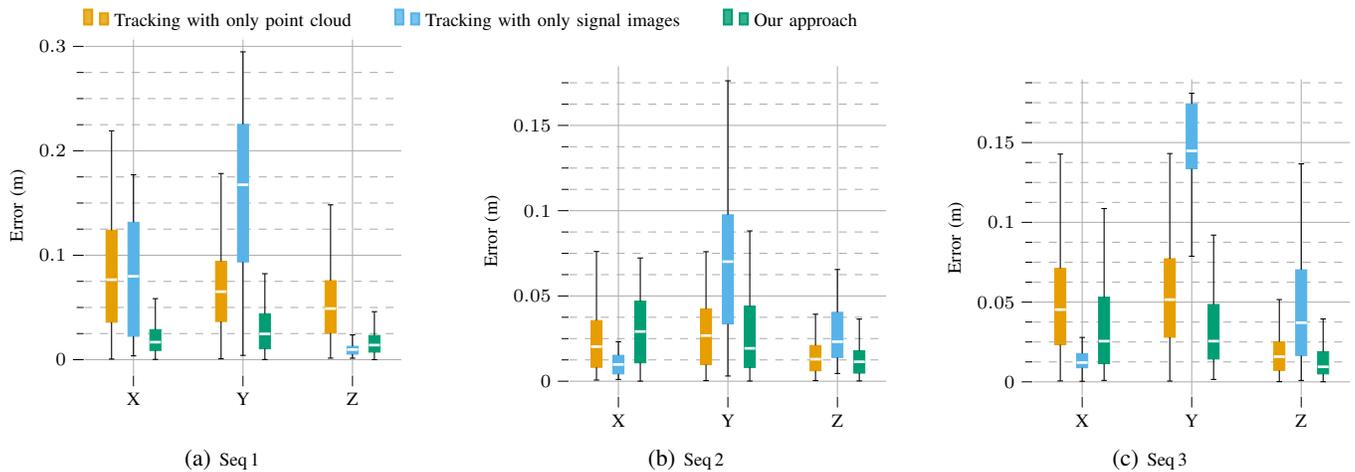

\begin{figure*}[t]
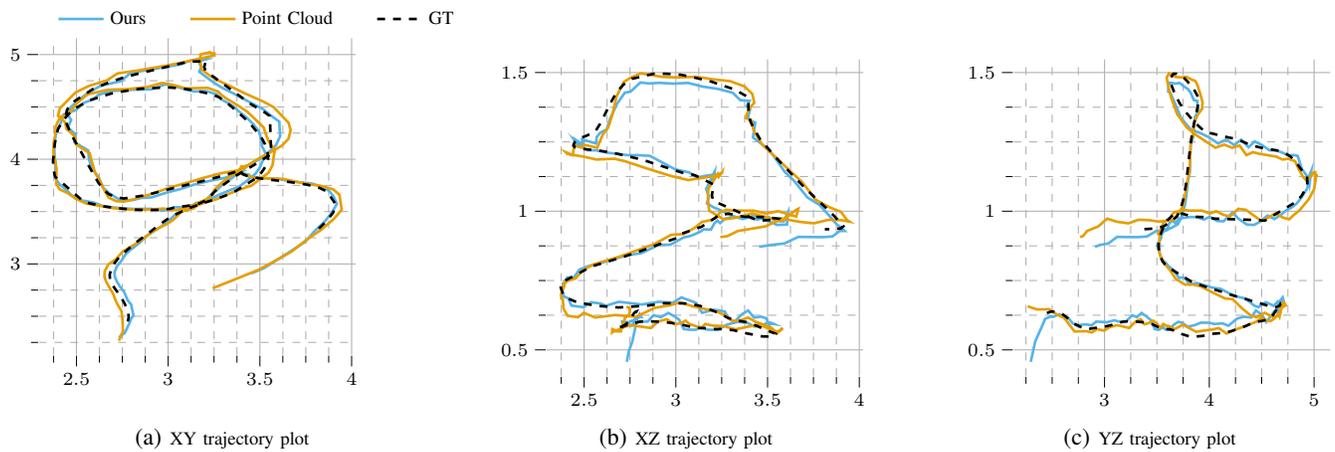

    \begin{subfigure}{.32\textwidth}
        \centering
        \setlength{\figurewidth}{\textwidth}
        \setlength{\figureheight}{\textwidth}
        \scriptsize{\input{tex/XYZ_plot/XY_plot_seq3}}
        \caption{\scriptsize{XY trajectory plot}}
        \label{fig:xy_plot}
    \end{subfigure}
    \hfill
    \begin{subfigure}{.32\textwidth}
        \centering
        \setlength{\figurewidth}{\textwidth}
        \setlength{\figureheight}{\textwidth}
        \scriptsize{\input{tex/XYZ_plot/XZ_plot_seq3}}
        \caption{\scriptsize{XZ trajectory plot}}
        \label{fig:xz_plot}     
    \end{subfigure}
    \hfill
    \begin{subfigure}{.32\textwidth}
        \centering
        \setlength{\figurewidth}{\textwidth}
        \setlength{\figureheight}{\textwidth}
        \scriptsize{\input{tex/XYZ_plot/YZ_plot_seq3}}
        \caption{\scriptsize{YZ trajectory plot}}
        \label{fig:yz_plot}     
    \end{subfigure}
    \caption{Comparison of estimated trajectories with the point cloud tracking method and our proposed method from three different projections.}
    \label{fig:full_traj} 
\end{figure*}

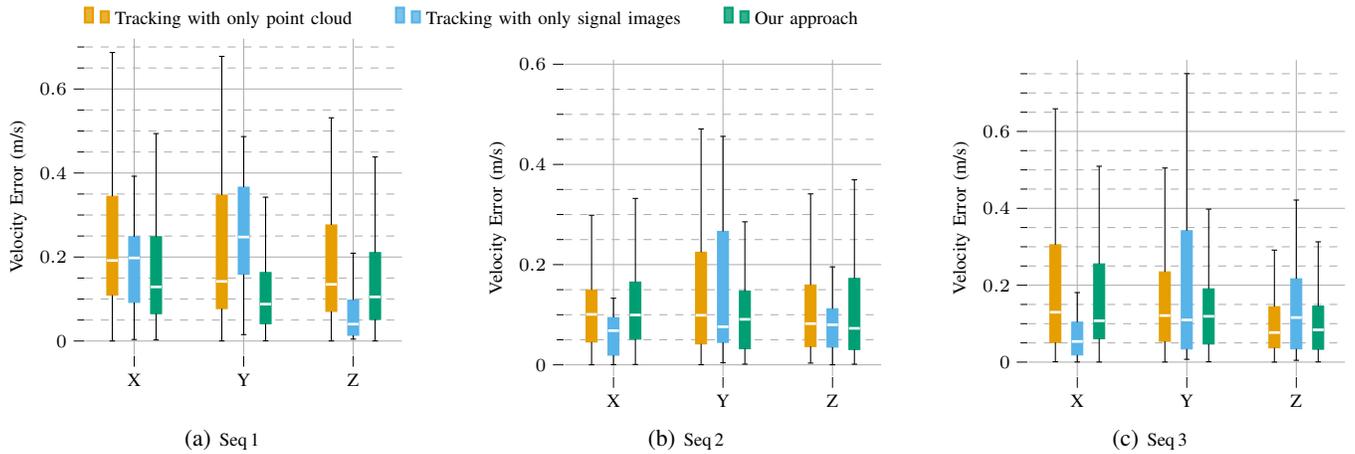
\begin{figure*}[t]
    \begin{subfigure}{.32\textwidth}
        \centering
        \setlength{\figurewidth}{\textwidth}
        \setlength{\figureheight}{\textwidth}
        \scriptsize{% This file was created with tikzplotlib v0.9.14.
\begin{tikzpicture}

% \definecolor{color0}{rgb}{0.6,0,0.980392156862745}
% \definecolor{color1}{rgb}{0.968627450980392,0.00784313725490196,0.164705882352941}
% \definecolor{color2}{rgb}{0.0235294117647059,0.32156862745098,1}
\definecolor{color0}{rgb}{0.90, 0.62, 0.00}
\definecolor{color1}{rgb}{0.34, 0.70, 0.91}
\definecolor{color2}{rgb}{0.00, 0.62, 0.45}
\definecolor{color3}{rgb}{0.94, 0.89, 0.27}
\definecolor{color4}{rgb}{0.00, 0.45, 0.69}
\definecolor{color5}{rgb}{0.83, 0.37, 0.00}

\begin{axis}[
    width=\figurewidth,
    height=\figureheight,
legend cell align={left},
legend style={
  fill opacity=0.8,
  draw opacity=1,
  text opacity=1,
  at={(-0.03,1.12)},
  anchor=north west,
  draw=white,
/tikz/every even column/.append style={column sep=0.5cm},
},
legend columns = 3, 
axis line style={white},
tick align=outside,
tick pos=left,
x grid style={white!69.0196078431373!black},
xmin=0.5, xmax=6.3,
xtick style={color=black},
xtick={1.4,3.4,5.4},
xticklabels={X,Y,Z},
y grid style={white!69.0196078431373!black},
ylabel={Velocity Error (m/s)},
ymin=-0.034208951695738, ymax=0.721221921251871,
ytick style={color=black},
    xmajorgrids,
    % xmajorticks=true,
    % xmin=-0.5, xmax=3.3125,
    xminorgrids,
    ymajorgrids,
    ymajorticks=true,
    minor y tick num = 3,
    minor y grid style={dashed},
    yminorgrids,
    yticklabel style={
            /pgf/number format/fixed,
            /pgf/number format/precision=5
    },
    scaled y ticks=false,
]

\draw[draw=color0,fill=color0, thick] (axis cs:0,0) rectangle (axis cs:0,0);
\addlegendimage{ybar, ybar legend, draw=color0,fill=color0,thick}
\addlegendentry{Tracking with only point cloud}

\draw[draw=color1,fill=color1, thick] (axis cs:0,0) rectangle (axis cs:0,0);
\addlegendimage{ybar, ybar legend, draw=color1,fill=color1,thick}
\addlegendentry{Tracking with only signal images}

\draw[draw=color2,fill=color2, thick] (axis cs:0,0) rectangle (axis cs:0,0);
\addlegendimage{ybar, ybar legend, draw=color2,fill=color2,thick}
\addlegendentry{Our approach}

\path [draw=color0, fill=color0]
(axis cs:0.9,0.109162671124885)
--(axis cs:1.1,0.109162671124885)
--(axis cs:1.1,0.344490432716165)
--(axis cs:0.9,0.344490432716165)
--(axis cs:0.9,0.109162671124885)
--cycle;
\addplot [black, forget plot]
table {%
1 0.109162671124885
1 0.000128815256426051
};
\addplot [black, forget plot]
table {%
1 0.344490432716165
1 0.686884154299707
};
\addplot [black, forget plot]
table {%
0.95 0.000128815256426051
1.05 0.000128815256426051
};
\addplot [black, forget plot]
table {%
0.95 0.686884154299707
1.05 0.686884154299707
};
\path [draw=color1, fill=color1]
(axis cs:1.3,0.0919291443556336)
--(axis cs:1.5,0.0919291443556336)
--(axis cs:1.5,0.24845085936301)
--(axis cs:1.3,0.24845085936301)
--(axis cs:1.3,0.0919291443556336)
--cycle;
\addplot [black, forget plot]
table {%
1.4 0.0919291443556336
1.4 0.00361160361343593
};
\addplot [black, forget plot]
table {%
1.4 0.24845085936301
1.4 0.392370148027017
};
\addplot [black, forget plot]
table {%
1.35 0.00361160361343593
1.45 0.00361160361343593
};
\addplot [black, forget plot]
table {%
1.35 0.392370148027017
1.45 0.392370148027017
};
\path [draw=color2, fill=color2]
(axis cs:1.7,0.0651457039411474)
--(axis cs:1.9,0.0651457039411474)
--(axis cs:1.9,0.248436061039956)
--(axis cs:1.7,0.248436061039956)
--(axis cs:1.7,0.0651457039411474)
--cycle;
\addplot [black, forget plot]
table {%
1.8 0.0651457039411474
1.8 0.0027373277432563
};
\addplot [black, forget plot]
table {%
1.8 0.248436061039956
1.8 0.493909091734377
};
\addplot [black, forget plot]
table {%
1.75 0.0027373277432563
1.85 0.0027373277432563
};
\addplot [black, forget plot]
table {%
1.75 0.493909091734377
1.85 0.493909091734377
};
\path [draw=color0, fill=color0]
(axis cs:2.9,0.0771871383898404)
--(axis cs:3.1,0.0771871383898404)
--(axis cs:3.1,0.347491246148062)
--(axis cs:2.9,0.347491246148062)
--(axis cs:2.9,0.0771871383898404)
--cycle;
\addplot [black, forget plot]
table {%
3 0.0771871383898404
3 0.000199890722223373
};
\addplot [black, forget plot]
table {%
3 0.347491246148062
3 0.677734254757247
};
\addplot [black, forget plot]
table {%
2.95 0.000199890722223373
3.05 0.000199890722223373
};
\addplot [black, forget plot]
table {%
2.95 0.677734254757247
3.05 0.677734254757247
};
\path [draw=color1, fill=color1]
(axis cs:3.3,0.158876445092808)
--(axis cs:3.5,0.158876445092808)
--(axis cs:3.5,0.366330405403881)
--(axis cs:3.3,0.366330405403881)
--(axis cs:3.3,0.158876445092808)
--cycle;
\addplot [black, forget plot]
table {%
3.4 0.158876445092808
3.4 0.0153142890715063
};
\addplot [black, forget plot]
table {%
3.4 0.366330405403881
3.4 0.486702403844719
};
\addplot [black, forget plot]
table {%
3.35 0.0153142890715063
3.45 0.0153142890715063
};
\addplot [black, forget plot]
table {%
3.35 0.486702403844719
3.45 0.486702403844719
};
\path [draw=color2, fill=color2]
(axis cs:3.7,0.0411681137598663)
--(axis cs:3.9,0.0411681137598663)
--(axis cs:3.9,0.163136180242405)
--(axis cs:3.7,0.163136180242405)
--(axis cs:3.7,0.0411681137598663)
--cycle;
\addplot [black, forget plot]
table {%
3.8 0.0411681137598663
3.8 0.000443114872510364
};
\addplot [black, forget plot]
table {%
3.8 0.163136180242405
3.8 0.342459269664861
};
\addplot [black, forget plot]
table {%
3.75 0.000443114872510364
3.85 0.000443114872510364
};
\addplot [black, forget plot]
table {%
3.75 0.342459269664861
3.85 0.342459269664861
};
\path [draw=color0, fill=color0]
(axis cs:4.9,0.070570502146341)
--(axis cs:5.1,0.070570502146341)
--(axis cs:5.1,0.276398234476372)
--(axis cs:4.9,0.276398234476372)
--(axis cs:4.9,0.070570502146341)
--cycle;
\addplot [black, forget plot]
table {%
5 0.070570502146341
5 0.000259953570465044
};
\addplot [black, forget plot]
table {%
5 0.276398234476372
5 0.531454865660448
};
\addplot [black, forget plot]
table {%
4.95 0.000259953570465044
5.05 0.000259953570465044
};
\addplot [black, forget plot]
table {%
4.95 0.531454865660448
5.05 0.531454865660448
};
\path [draw=color1, fill=color1]
(axis cs:5.3,0.0136535369876245)
--(axis cs:5.5,0.0136535369876245)
--(axis cs:5.5,0.0975762615115577)
--(axis cs:5.3,0.0975762615115577)
--(axis cs:5.3,0.0136535369876245)
--cycle;
\addplot [black, forget plot]
table {%
5.4 0.0136535369876245
5.4 0.00456348111469773
};
\addplot [black, forget plot]
table {%
5.4 0.0975762615115577
5.4 0.209099309075048
};
\addplot [black, forget plot]
table {%
5.35 0.00456348111469773
5.45 0.00456348111469773
};
\addplot [black, forget plot]
table {%
5.35 0.209099309075048
5.45 0.209099309075048
};
\path [draw=color2, fill=color2]
(axis cs:5.7,0.051383654669091)
--(axis cs:5.9,0.051383654669091)
--(axis cs:5.9,0.210674396423683)
--(axis cs:5.7,0.210674396423683)
--(axis cs:5.7,0.051383654669091)
--cycle;
\addplot [black, forget plot]
table {%
5.8 0.051383654669091
5.8 0.000249636769975581
};
\addplot [black, forget plot]
table {%
5.8 0.210674396423683
5.8 0.438200065621051
};
\addplot [black, forget plot]
table {%
5.75 0.000249636769975581
5.85 0.000249636769975581
};
\addplot [black, forget plot]
table {%
5.75 0.438200065621051
5.85 0.438200065621051
};
\addplot [line width=1pt, white, forget plot]
table {%
0.9 0.191737841303157
1.1 0.191737841303157
};
\addplot [line width=1pt, white, forget plot]
table {%
1.3 0.197800169460929
1.5 0.197800169460929
};
\addplot [line width=1pt, white, forget plot]
table {%
1.7 0.128881545706658
1.9 0.128881545706658
};
\addplot [line width=1pt, white, forget plot]
table {%
2.9 0.141984442865786
3.1 0.141984442865786
};
\addplot [line width=1pt, white, forget plot]
table {%
3.3 0.247618886526242
3.5 0.247618886526242
};
\addplot [line width=1pt, white, forget plot]
table {%
3.7 0.0878626303248753
3.9 0.0878626303248753
};
\addplot [line width=1pt, white, forget plot]
table {%
4.9 0.134968087815255
5.1 0.134968087815255
};
\addplot [line width=1pt, white, forget plot]
table {%
5.3 0.0402095843975281
5.5 0.0402095843975281
};
\addplot [line width=1pt, white, forget plot]
table {%
5.7 0.105071468625511
5.9 0.105071468625511
};
\end{axis}

\end{tikzpicture}}
        \caption{\scriptsize{Seq\,1}}
        \label{fig:vel_err_seq1}
    \end{subfigure}
    \hfill
    \begin{subfigure}{.32\textwidth}
        \centering
        \setlength{\figurewidth}{\textwidth}
        \setlength{\figureheight}{\textwidth}
        \scriptsize{% This file was created with tikzplotlib v0.9.14.
\begin{tikzpicture}

% \definecolor{color0}{rgb}{0.6,0,0.980392156862745}
% \definecolor{color1}{rgb}{0.968627450980392,0.00784313725490196,0.164705882352941}
% \definecolor{color2}{rgb}{0.0235294117647059,0.32156862745098,1}
\definecolor{color0}{rgb}{0.90, 0.62, 0.00}
\definecolor{color1}{rgb}{0.34, 0.70, 0.91}
\definecolor{color2}{rgb}{0.00, 0.62, 0.45}
\definecolor{color3}{rgb}{0.94, 0.89, 0.27}
\definecolor{color4}{rgb}{0.00, 0.45, 0.69}
\definecolor{color5}{rgb}{0.83, 0.37, 0.00}

\begin{axis}[
    width=\figurewidth,
    height=\figureheight,
legend cell align={left},
legend style={
  fill opacity=0.8,
  draw opacity=1,
  text opacity=1,
  at={(0.03,0.97)},
  anchor=north west,
  draw=white!80!black
},
axis line style={white},
tick align=outside,
tick pos=left,
x grid style={white!69.0196078431373!black},
xmin=0.5, xmax=6.3,
xtick style={color=black},
xtick={1.4,3.4,5.4},
xticklabels={X,Y,Z},
y grid style={white!69.0196078431373!black},
ylabel={Velocity Error (m/s)},
ymin=-0.0233556679147973, ymax=0.61,
ytick style={color=black},
    xmajorgrids,
    % xmajorticks=true,
    % xmin=-0.5, xmax=3.3125,
    xminorgrids,
    ymajorgrids,
    ymajorticks=true,
    minor y tick num = 3,
    minor y grid style={dashed},
    yminorgrids,
    yticklabel style={
            /pgf/number format/fixed,
            /pgf/number format/precision=5
    },
    scaled y ticks=false,
]
\path [draw=color0, fill=color0]
(axis cs:0.9,0.0454993116751612)
--(axis cs:1.1,0.0454993116751612)
--(axis cs:1.1,0.149575866732583)
--(axis cs:0.9,0.149575866732583)
--(axis cs:0.9,0.0454993116751612)
--cycle;
\addplot [black, forget plot]
table {%
1 0.0454993116751612
1 0.000176190755700745
};
\addplot [black, forget plot]
table {%
1 0.149575866732583
1 0.298551854208258
};
\addplot [black, forget plot]
table {%
0.95 0.000176190755700745
1.05 0.000176190755700745
};
\addplot [black, forget plot]
table {%
0.95 0.298551854208258
1.05 0.298551854208258
};
\path [draw=color1, fill=color1]
(axis cs:1.3,0.0193088946071396)
--(axis cs:1.5,0.0193088946071396)
--(axis cs:1.5,0.0943910742063947)
--(axis cs:1.3,0.0943910742063947)
--(axis cs:1.3,0.0193088946071396)
--cycle;
\addplot [black, forget plot]
table {%
1.4 0.0193088946071396
1.4 0.000377116058847271
};
\addplot [black, forget plot]
table {%
1.4 0.0943910742063947
1.4 0.132937731845124
};
\addplot [black, forget plot]
table {%
1.35 0.000377116058847271
1.45 0.000377116058847271
};
\addplot [black, forget plot]
table {%
1.35 0.132937731845124
1.45 0.132937731845124
};
\path [draw=color2, fill=color2]
(axis cs:1.7,0.0512629607750226)
--(axis cs:1.9,0.0512629607750226)
--(axis cs:1.9,0.165225579088091)
--(axis cs:1.7,0.165225579088091)
--(axis cs:1.7,0.0512629607750226)
--cycle;
\addplot [black, forget plot]
table {%
1.8 0.0512629607750226
1.8 0.00075650953687223
};
\addplot [black, forget plot]
table {%
1.8 0.165225579088091
1.8 0.332121025821897
};
\addplot [black, forget plot]
table {%
1.75 0.00075650953687223
1.85 0.00075650953687223
};
\addplot [black, forget plot]
table {%
1.75 0.332121025821897
1.85 0.332121025821897
};
\path [draw=color0, fill=color0]
(axis cs:2.9,0.0417637473191679)
--(axis cs:3.1,0.0417637473191679)
--(axis cs:3.1,0.224785137712553)
--(axis cs:2.9,0.224785137712553)
--(axis cs:2.9,0.0417637473191679)
--cycle;
\addplot [black, forget plot]
table {%
3 0.0417637473191679
3 0.000264292467084815
};
\addplot [black, forget plot]
table {%
3 0.224785137712553
3 0.470813364165661
};
\addplot [black, forget plot]
table {%
2.95 0.000264292467084815
3.05 0.000264292467084815
};
\addplot [black, forget plot]
table {%
2.95 0.470813364165661
3.05 0.470813364165661
};
\path [draw=color1, fill=color1]
(axis cs:3.3,0.0448127976552481)
--(axis cs:3.5,0.0448127976552481)
--(axis cs:3.5,0.266378868864894)
--(axis cs:3.3,0.266378868864894)
--(axis cs:3.3,0.0448127976552481)
--cycle;
\addplot [black, forget plot]
table {%
3.4 0.0448127976552481
3.4 0.00411012798469557
};
\addplot [black, forget plot]
table {%
3.4 0.266378868864894
3.4 0.456005566679127
};
\addplot [black, forget plot]
table {%
3.35 0.00411012798469557
3.45 0.00411012798469557
};
\addplot [black, forget plot]
table {%
3.35 0.456005566679127
3.45 0.456005566679127
};
\path [draw=color2, fill=color2]
(axis cs:3.7,0.032206746853467)
--(axis cs:3.9,0.032206746853467)
--(axis cs:3.9,0.147270781441821)
--(axis cs:3.7,0.147270781441821)
--(axis cs:3.7,0.032206746853467)
--cycle;
\addplot [black, forget plot]
table {%
3.8 0.032206746853467
3.8 0.00140957279476162
};
\addplot [black, forget plot]
table {%
3.8 0.147270781441821
3.8 0.285432958940515
};
\addplot [black, forget plot]
table {%
3.75 0.00140957279476162
3.85 0.00140957279476162
};
\addplot [black, forget plot]
table {%
3.75 0.285432958940515
3.85 0.285432958940515
};
\path [draw=color0, fill=color0]
(axis cs:4.9,0.0365176475868445)
--(axis cs:5.1,0.0365176475868445)
--(axis cs:5.1,0.159425725571298)
--(axis cs:4.9,0.159425725571298)
--(axis cs:4.9,0.0365176475868445)
--cycle;
\addplot [black, forget plot]
table {%
5 0.0365176475868445
5 0.00350969869741924
};
\addplot [black, forget plot]
table {%
5 0.159425725571298
5 0.341189883516642
};
\addplot [black, forget plot]
table {%
4.95 0.00350969869741924
5.05 0.00350969869741924
};
\addplot [black, forget plot]
table {%
4.95 0.341189883516642
5.05 0.341189883516642
};
\path [draw=color1, fill=color1]
(axis cs:5.3,0.0352932972645825)
--(axis cs:5.5,0.0352932972645825)
--(axis cs:5.5,0.111943745306393)
--(axis cs:5.3,0.111943745306393)
--(axis cs:5.3,0.0352932972645825)
--cycle;
\addplot [black, forget plot]
table {%
5.4 0.0352932972645825
5.4 0.000227750891903522
};
\addplot [black, forget plot]
table {%
5.4 0.111943745306393
5.4 0.195521128516952
};
\addplot [black, forget plot]
table {%
5.35 0.000227750891903522
5.45 0.000227750891903522
};
\addplot [black, forget plot]
table {%
5.35 0.195521128516952
5.45 0.195521128516952
};
\path [draw=color2, fill=color2]
(axis cs:5.7,0.0305767420525668)
--(axis cs:5.9,0.0305767420525668)
--(axis cs:5.9,0.172522100116694)
--(axis cs:5.7,0.172522100116694)
--(axis cs:5.7,0.0305767420525668)
--cycle;
\addplot [black, forget plot]
table {%
5.8 0.0305767420525668
5.8 0.00128623420413865
};
\addplot [black, forget plot]
table {%
5.8 0.172522100116694
5.8 0.369336006006802
};
\addplot [black, forget plot]
table {%
5.75 0.00128623420413865
5.85 0.00128623420413865
};
\addplot [black, forget plot]
table {%
5.75 0.369336006006802
5.85 0.369336006006802
};
\addplot [line width=1pt, white, forget plot]
table {%
0.9 0.100766256788645
1.1 0.100766256788645
};
\addplot [line width=1pt, white, forget plot]
table {%
1.3 0.0682491611946423
1.5 0.0682491611946423
};
\addplot [line width=1pt, white, forget plot]
table {%
1.7 0.099464031339338
1.9 0.099464031339338
};
\addplot [line width=1pt, white, forget plot]
table {%
2.9 0.0989566119088892
3.1 0.0989566119088892
};
\addplot [line width=1pt, white, forget plot]
table {%
3.3 0.075984746026001
3.5 0.075984746026001
};
\addplot [line width=1pt, white, forget plot]
table {%
3.7 0.0908502564243308
3.9 0.0908502564243308
};
\addplot [line width=1pt, white, forget plot]
table {%
4.9 0.0821778973760645
5.1 0.0821778973760645
};
\addplot [line width=1pt, white, forget plot]
table {%
5.3 0.079976425217278
5.5 0.079976425217278
};
\addplot [line width=1pt, white, forget plot]
table {%
5.7 0.0728472717659245
5.9 0.0728472717659245
};
\end{axis}

\end{tikzpicture}}
        \caption{\scriptsize{Seq\,2}}
        \label{fig:vel_err_seq2}     
    \end{subfigure}
    \hfill
    \begin{subfigure}{.32\textwidth}
        \centering
        \setlength{\figurewidth}{\textwidth}
        \setlength{\figureheight}{\textwidth}
        \scriptsize{% This file was created with tikzplotlib v0.9.14.
\begin{tikzpicture}

% \definecolor{color0}{rgb}{0.6,0,0.980392156862745}
% \definecolor{color1}{rgb}{0.968627450980392,0.00784313725490196,0.164705882352941}
% \definecolor{color2}{rgb}{0.0235294117647059,0.32156862745098,1}
\definecolor{color0}{rgb}{0.90, 0.62, 0.00}
\definecolor{color1}{rgb}{0.34, 0.70, 0.91}
\definecolor{color2}{rgb}{0.00, 0.62, 0.45}
\definecolor{color3}{rgb}{0.94, 0.89, 0.27}
\definecolor{color4}{rgb}{0.00, 0.45, 0.69}
\definecolor{color5}{rgb}{0.83, 0.37, 0.00}

\begin{axis}[
    width=\figurewidth,
    height=\figureheight,
legend cell align={left},
legend style={
  fill opacity=0.8,
  draw opacity=1,
  text opacity=1,
  at={(0.03,0.97)},
  anchor=north west,
  draw=white!80!black
},
axis line style={white},
tick align=outside,
tick pos=left,
x grid style={white!69.0196078431373!black},
xmin=0.5, xmax=6.3,
xtick style={color=black},
xtick={1.4,3.4,5.4},
xticklabels={X,Y,Z},
y grid style={white!69.0196078431373!black},
ylabel={Velocity Error (m/s)},
ymin=-0.0375045054404742, ymax=0.78781873238659,
ytick style={color=black},
    xmajorgrids,
    % xmajorticks=true,
    % xmin=-0.5, xmax=3.3125,
    xminorgrids,
    ymajorgrids,
    ymajorticks=true,
    minor y tick num = 3,
    minor y grid style={dashed},
    yminorgrids,
    yticklabel style={
            /pgf/number format/fixed,
            /pgf/number format/precision=5
    },
    scaled y ticks=false,
]
\path [draw=color0, fill=color0]
(axis cs:0.9,0.050925850163972)
--(axis cs:1.1,0.050925850163972)
--(axis cs:1.1,0.305209551337428)
--(axis cs:0.9,0.305209551337428)
--(axis cs:0.9,0.050925850163972)
--cycle;
\addplot [black, forget plot]
table {%
1 0.050925850163972
1 0.000915280434687915
};
\addplot [black, forget plot]
table {%
1 0.305209551337428
1 0.658851716939473
};
\addplot [black, forget plot]
table {%
0.95 0.000915280434687915
1.05 0.000915280434687915
};
\addplot [black, forget plot]
table {%
0.95 0.658851716939473
1.05 0.658851716939473
};
\path [draw=color1, fill=color1]
(axis cs:1.3,0.0186842481424454)
--(axis cs:1.5,0.0186842481424454)
--(axis cs:1.5,0.103976060431693)
--(axis cs:1.3,0.103976060431693)
--(axis cs:1.3,0.0186842481424454)
--cycle;
\addplot [black, forget plot]
table {%
1.4 0.0186842481424454
1.4 0.000318754243475716
};
\addplot [black, forget plot]
table {%
1.4 0.103976060431693
1.4 0.180866270180915
};
\addplot [black, forget plot]
table {%
1.35 0.000318754243475716
1.45 0.000318754243475716
};
\addplot [black, forget plot]
table {%
1.35 0.180866270180915
1.45 0.180866270180915
};
\path [draw=color2, fill=color2]
(axis cs:1.7,0.0607938102103689)
--(axis cs:1.9,0.0607938102103689)
--(axis cs:1.9,0.255498452503433)
--(axis cs:1.7,0.255498452503433)
--(axis cs:1.7,0.0607938102103689)
--cycle;
\addplot [black, forget plot]
table {%
1.8 0.0607938102103689
1.8 0.000215207845459808
};
\addplot [black, forget plot]
table {%
1.8 0.255498452503433
1.8 0.509395351885571
};
\addplot [black, forget plot]
table {%
1.75 0.000215207845459808
1.85 0.000215207845459808
};
\addplot [black, forget plot]
table {%
1.75 0.509395351885571
1.85 0.509395351885571
};
\path [draw=color0, fill=color0]
(axis cs:2.9,0.0540813292258191)
--(axis cs:3.1,0.0540813292258191)
--(axis cs:3.1,0.234681439454613)
--(axis cs:2.9,0.234681439454613)
--(axis cs:2.9,0.0540813292258191)
--cycle;
\addplot [black, forget plot]
table {%
3 0.0540813292258191
3 0.00025457195519607
};
\addplot [black, forget plot]
table {%
3 0.234681439454613
3 0.50483313777927
};
\addplot [black, forget plot]
table {%
2.95 0.00025457195519607
3.05 0.00025457195519607
};
\addplot [black, forget plot]
table {%
2.95 0.50483313777927
3.05 0.50483313777927
};
\path [draw=color1, fill=color1]
(axis cs:3.3,0.0341207415404121)
--(axis cs:3.5,0.0341207415404121)
--(axis cs:3.5,0.341801049261369)
--(axis cs:3.3,0.341801049261369)
--(axis cs:3.3,0.0341207415404121)
--cycle;
\addplot [black, forget plot]
table {%
3.4 0.0341207415404121
3.4 0.00729099756574314
};
\addplot [black, forget plot]
table {%
3.4 0.341801049261369
3.4 0.750304039758087
};
\addplot [black, forget plot]
table {%
3.35 0.00729099756574314
3.45 0.00729099756574314
};
\addplot [black, forget plot]
table {%
3.35 0.750304039758087
3.45 0.750304039758087
};
\path [draw=color2, fill=color2]
(axis cs:3.7,0.0470434830938093)
--(axis cs:3.9,0.0470434830938093)
--(axis cs:3.9,0.190534730347155)
--(axis cs:3.7,0.190534730347155)
--(axis cs:3.7,0.0470434830938093)
--cycle;
\addplot [black, forget plot]
table {%
3.8 0.0470434830938093
3.8 0.000923772653984578
};
\addplot [black, forget plot]
table {%
3.8 0.190534730347155
3.8 0.397895565899473
};
\addplot [black, forget plot]
table {%
3.75 0.000923772653984578
3.85 0.000923772653984578
};
\addplot [black, forget plot]
table {%
3.75 0.397895565899473
3.85 0.397895565899473
};
\path [draw=color0, fill=color0]
(axis cs:4.9,0.0370850724055838)
--(axis cs:5.1,0.0370850724055838)
--(axis cs:5.1,0.144242538996679)
--(axis cs:4.9,0.144242538996679)
--(axis cs:4.9,0.0370850724055838)
--cycle;
\addplot [black, forget plot]
table {%
5 0.0370850724055838
5 1.01871880286986e-05
};
\addplot [black, forget plot]
table {%
5 0.144242538996679
5 0.29127738534024
};
\addplot [black, forget plot]
table {%
4.95 1.01871880286986e-05
5.05 1.01871880286986e-05
};
\addplot [black, forget plot]
table {%
4.95 0.29127738534024
5.05 0.29127738534024
};
\path [draw=color1, fill=color1]
(axis cs:5.3,0.0345189067087442)
--(axis cs:5.5,0.0345189067087442)
--(axis cs:5.5,0.216801981745139)
--(axis cs:5.3,0.216801981745139)
--(axis cs:5.3,0.0345189067087442)
--cycle;
\addplot [black, forget plot]
table {%
5.4 0.0345189067087442
5.4 0.00468124038173312
};
\addplot [black, forget plot]
table {%
5.4 0.216801981745139
5.4 0.421482532405152
};
\addplot [black, forget plot]
table {%
5.35 0.00468124038173312
5.45 0.00468124038173312
};
\addplot [black, forget plot]
table {%
5.35 0.421482532405152
5.45 0.421482532405152
};
\path [draw=color2, fill=color2]
(axis cs:5.7,0.0333654972375574)
--(axis cs:5.9,0.0333654972375574)
--(axis cs:5.9,0.145628655344953)
--(axis cs:5.7,0.145628655344953)
--(axis cs:5.7,0.0333654972375574)
--cycle;
\addplot [black, forget plot]
table {%
5.8 0.0333654972375574
5.8 0.000549304948028739
};
\addplot [black, forget plot]
table {%
5.8 0.145628655344953
5.8 0.312894917001754
};
\addplot [black, forget plot]
table {%
5.75 0.000549304948028739
5.85 0.000549304948028739
};
\addplot [black, forget plot]
table {%
5.75 0.312894917001754
5.85 0.312894917001754
};
\addplot [line width=1pt, white, forget plot]
table {%
0.9 0.129738371745527
1.1 0.129738371745527
};
\addplot [line width=1pt, white, forget plot]
table {%
1.3 0.0536968219671419
1.5 0.0536968219671419
};
\addplot [line width=1pt, white, forget plot]
table {%
1.7 0.107509878653638
1.9 0.107509878653638
};
\addplot [line width=1pt, white, forget plot]
table {%
2.9 0.121060592272317
3.1 0.121060592272317
};
\addplot [line width=1pt, white, forget plot]
table {%
3.3 0.109664642493887
3.5 0.109664642493887
};
\addplot [line width=1pt, white, forget plot]
table {%
3.7 0.119131419941609
3.9 0.119131419941609
};
\addplot [line width=1pt, white, forget plot]
table {%
4.9 0.0768678127111805
5.1 0.0768678127111805
};
\addplot [line width=1pt, white, forget plot]
table {%
5.3 0.115904482344118
5.5 0.115904482344118
};
\addplot [line width=1pt, white, forget plot]
table {%
5.7 0.083997795576003
5.9 0.083997795576003
};
\end{axis}

\end{tikzpicture}}
        \caption{\scriptsize{Seq\,3}}
        \label{fig:vel_err_seq3}     
    \end{subfigure}
    \caption{Velocity estimation error for each linear component in the three data sequences.}
    \label{fig:vel_errs} 
\end{figure*}

\section{Experimental Results}

In this section, we report the experimental results, based on the three data sequences gathered in the indoor test environment. 

\subsection{UAV in the Ouster LiDAR point cloud}

The first parameter to analyze is the number of points that reflect from the UAV at different distances. Our analysis, shown in Fig~\ref{fig:pcd_frame}, reveals that the point cloud structure generated by the UAV is significantly influenced by the distance from the target. At short distances, the point cloud produced by LiDAR is abundant and presents comprehensive details. When the distance extends to a medium range, the number of UAV point clouds decreases to less than 100, but the three-dimensional structure of the UAV remains discernible. However, when the distance is at a medium range of 7\,m according to our results, the number of UAV point clouds reduces to single digits, and the point cloud structure becomes highly unpredictable and unstructured. It is worth noting that in a more realistic application, additional elements such as other sensor payloads or a cargo bay would potentially increase significantly the reflective surface of the UAV.

\subsection{Trajectory Validation}

We also show a quantitative analysis of the APE based on ground truth, with the main results summarized in Fig.~\ref{fig:ape_error}. To ensure the trajectories are compared under the same coordinates, we utilize the coordinates of the ground truth as the reference coordinates and convert all trajectories generated by the three UAV tracking methods to these coordinates. 
% For the primary evaluation metric, we employ the absolute pose error (APE) [29]. To compute the error for each trajectory, we use the open-source EVO toolset. This enables us to conduct a comprehensive evaluation of the accuracy of each tracking method.
Table~\ref{tab:methods_compare} presents a comprehensive comparison of three different UAV tracking methods in terms of detectable distance, average APE, algorithm update frequency, and need for initial conditions. 
\begin{table}[h]
    \centering
    \caption{ Performance( Detectable distance, frame rate, and APE error) and initial conditions comparison of selected tracking methods. } 
    \resizebox{0.48\textwidth}{!}{%
    \begin{tabular}{@{}lcccc@{}}
        \toprule
        & \textbf{Distance} & \textbf{APE Error(Mean/RMSE)} & \textbf{FPS} & \textbf{ Initialization} \\ 
        & \textit{(m)} & (m) & (Hz) \\
        \midrule
        \textbf{Tracking with point cloud} & \textbf{8.0}  & 0.104 / 0.142 & 8.3  & Yes  \\
        \textbf{Tracking with signal images} & 2.4  & 0.078 / 0.088 & \textbf{10} & No \\
        \textbf{Fused(Ours)} & \textbf{8.0}  & \textbf{0.061 / 0.067} & \textbf{10} & No \\
        \bottomrule
    \end{tabular}
    }
    \label{tab:methods_compare}
\end{table}

The image-based UAV tracking method shows a relatively small average error; however, its overall error distribution is inconsistent, as the Y-axis error in the \textit{Seq\,1} sequence reaches up to 0.3\,m. Conversely, the point cloud-based UAV tracking method has the largest average error, but its error distribution is more uniform. Our proposed method, on the other hand, achieves the smallest average error and minimal error fluctuation. Additionally, to supplement the quantitative trajectory analysis, we also provide a visualization of the trajectories based on the point cloud tracking method and our proposed method from three different viewpoints, as illustrated in Fig~\ref{fig:full_traj}, with more consistent behavior.

\subsection{Velocity Validation}

In addition to pose estimation, we conducted a quantitative analysis of the UAV velocities based on the ground truth data and compared them with different UAV tracking methods. Fig.~\ref{fig:vel_errs} illustrates the velocity errors of each method along the X, Y, and Z axes. The experimental results reveal that all methods have similar mean values of the velocity errors, but different fluctuations. The image tracking method has a large fluctuation in the Y-axis velocity error, reaching up to \textit{0.75 m/s} in \textit{Seq\,3}. The point cloud tracking method also has relatively large fluctuations in all dimensions. In contrast, our method achieves smaller overall velocity errors in both the mean value and fluctuation range.

\subsection{Resource Consumption}
% To validate the performance of our UAV tracking design and better understand its limitations, we collected three flight trajectory sequences (\textit{Seq 1}, \textit{Seq 2}, and \textit{Seq 3}) in a large open area of 10 x 10 square meters, 0.5 to 8 meters away from the LiDAR scanner, as detailed in Table~\ref{tab:sequence_detail}. \textit{Seq 1} and \textit{Seq 3} represent a helical ascension trajectory, while \textit{Seq 2} represents an elliptical trajectory. We first focused on analyzing \textit{Seq 3} and compared it to the ground truth trajectory obtained from the Mocap system. 

% We conducted the experiment on two different platforms, the Lenovo Legion Y7000P equipped with 16GB RAM, 6-core Intel i5-9300H (2.40GHz) and Nvidia GTX 1660Ti (1536 CUDA cores, 6GB VRAM), as well as the commonly used embedded computing platform Jetson Nano with 4-core ARM A57 64-bit CPU (1.43GHz), 4GB RAM, and 128-core Maxwell GPU. 
Both the Intel laptop and the Jetson Nano run ROS Melodic on Ubuntu 18.04. The CPU and memory utilization is measured with a ROS resource monitor tool~\footnote{\href{https://github.com/alspitz/cpu_monitor}{https://github.com/alspitz/cpu\_monitor}}. Additionally, for minimizing the difference of the operating environment, we unified the dependencies used in each method into same version. The results are summarized in Table~\ref{tab:runtime_src}.

The memory utilization of each selected method was roughly equivalent in both processor architecture platforms. However, the same algorithm showed generally higher CPU utilization and achieved the highest publishing rate when running on the Intel processor. For the Intel processor, the point cloud tracking method had higher CPU utilization than other methods but the lowest publishing rate. The fusion method performed well on the laptop and had the smallest overall error. On the embedded computing platform, the CPU utilization of all methods did not differ significantly, and the point cloud tracking method had the lowest memory utilization but the lowest pose publication rate. The difference in CPU utilization is caused by the use of CUDA GPU acceleration in the Open3D binaries utilized for the Jetson Nano platform, while the Intel computer uses only the CPU for point cloud data processing. The image processing also leverages the embedded GPU in the Jetson Nano board. Because of the small ROI that we extract to process the point cloud, the fused method adds little overhead on top of the vision-only method.

\begin{table}[t]
    \centering
    \caption{ Average run-time resource (CPU/RAM) utilization and performance (pose calculation speed) comparison of selected tracking methods across multiple platforms. CPU utilization of 100\% equals one full processor core. } 
    % \renewcommand{\arraystretch}{1.6}
    % \resizebox{\linewidth}{!}{%
    \begin{tabular}{@{}lcc@{}}
        \toprule
         & Laptop & Jetson Nano  \\
        & \multicolumn{2}{c}{\textit{( CPU (\%), RAM (MB), Pose rate (Hz) )}} \\%[-.21em]
        \midrule
        \textbf{Tracking with point cloud} & (422.7, 296.1, 8.3 ) & (121.7, 179.7, 5.15 )  \\
        \textbf{Tracking with image} & (209.0, 293.6, 10 ) & (113.8, 232.9, 6.07)   \\
        \textbf{Fused (ours)} & (195.5, 299.0, 10 ) & (114.5, 247.8, 6.04)    \\ \bottomrule
    \end{tabular}
    % }
    \label{tab:runtime_src}
\end{table}

\section{Conclusion}\label{sec:conclusion}
This paper has proposed a novel approach for tracking a UAV based on the fusion of signal images and point clouds from an Ouster LiDAR. Unlike conventional LiDAR and camera fusion, this approach does not need any calibration and preprocessing with external cameras and the LiDAR data is more resistant to harsh environments. We collected three different data sequences in an indoor environment with the OptiTrack mocap system providing ground truth positions. We compared the proposed approach with the approaches based on either only point clouds or signal images and the results showed the effectiveness of our proposed approach. Additionally, we found that our approach can be utilized in a popular mobile computing platform, Jetson Nano according to our evaluation.

Future work includes fusing the Ouster images (depth, signal, and ambient), point clouds, and conventional RGB images in various applications including simultaneous localization
and mapping (SLAM), object detection, and tracking.

% \newpage
%%%%%%%%%%%%%%%%%%%%%%%%%%%%%%%%%%%%%%%%%%%%%%
%%                                          %%
%%            ACKNOWLEDGMENT                %%
%%                                          %%
%%%%%%%%%%%%%%%%%%%%%%%%%%%%%%%%%%%%%%%%%%%%%%

\section*{Acknowledgment}

This research work is supported by the Academy of Finland's AeroPolis project (Grant No. 348480).

%%%%%%%%%%%%%%%%%%%%%%%%%%%%%%%%%%%%%%%%%%%%%%
%%                                          %%
%%              BIBLIOGRAPHY                %%
%%                                          %%
%%%%%%%%%%%%%%%%%%%%%%%%%%%%%%%%%%%%%%%%%%%%%%
% \newpage
% \nocite{*}
\bibliographystyle{unsrt}
\bibliography{bibliography}

\end{document}